\documentclass[letterpaper]{article} 
\usepackage[utf8]{inputenc}
\usepackage{aaai2027}
\usepackage{natbib}

\newlength{\colRole}
\newlength{\colContent}
\usepackage{times}
\usepackage{helvet}
\usepackage{courier}

\usepackage{graphicx}
\usepackage{amsmath}
\usepackage{array}
\usepackage{amssymb}
\usepackage{booktabs}
\usepackage{tabularx}
\usepackage{multirow}
\usepackage{url}
\usepackage{graphicx}
\usepackage{caption}
\usepackage{tabularx}
\usepackage[table]{xcolor}
\usepackage{enumitem}
\usepackage{pifont}
\usepackage{footmisc}

\renewcommand{\footnotesize}{\scriptsize}


\newcolumntype{J}[1]{>{\arraybackslash\hsize=#1\hsize}X}
\newcolumntype{L}{>{\raggedright\arraybackslash}X}
\newcolumntype{C}[1]{>{\centering\arraybackslash}p{#1}}

\newcommand{\rotlabel}[2][0mm]{%
  \raisebox{#1}{\rotatebox[origin=c]{90}{\textbf{#2}}}%
}

\title{ \textsc{Cognitive World Model} for Progressive BDI/E Trajectory Evaluation of Conversational Agents}

\author{
    Minghui Ma\textsuperscript{\rm 1},
    Bin Guo\textsuperscript{\rm 1},
    Hao Wang\textsuperscript{\rm 2} ,
    Han Wang ,
    Mengqi Chen\textsuperscript{\rm 1},
    Jingqi Liu\textsuperscript{\rm 1},
    Yan Liu\textsuperscript{\rm 1},
}

\affiliations{
    \textsuperscript{1} Northwestern Polytechnical University,
    \textsuperscript{2} Xi'an Jiaotong University, Xi'an, China
}
\begin{document}

\maketitle


\begin{abstract}
As LLM-based conversational agents advance toward increasingly open-ended and interaction-intensive scenarios, task completion alone provides an incomplete assessment of their effectiveness. The evolution of users' internal states, including beliefs, desires, intentions, and emotions (BDI/E), serves as an intermediate signal connecting agent behaviors with interaction outcomes and reflects how conversational strategies shape users during multi-turn interactions. However, existing evaluation paradigms primarily focus on surface-level responses or final outcomes, providing limited insight into the underlying cognitive processes. This limitation makes it difficult to diagnose why agents succeed or fail and to optimize their interaction strategies. To address this challenge, we propose \textbf{Cognitive World Model (CogWM)}, an LLM-based cognitive user model that jointly models users' BDI/E states and corresponding responses, enabling explicit cognitive trajectory tracking. Trained on 150K user-turn samples with Qwen3-14B, CogWM achieves superior performance over existing user simulation baselines in both response fidelity and cognitive state understanding. Interactions with six state-of-the-art LLMs demonstrate that CogWM enables progressive comparison of agents through cognitive trajectories, revealing distinct agent patterns and complementary relationships between cognitive evolution and behavioral outcomes. We have released our code\footnote{\scriptsize Code: https://github.com/lucianma05-create/CogWM} and models\footnote{Model: https://www.modelscope.cn/models/LucianMa/CogWM-14B}.
\end{abstract}

\section{Introduction}
\label{sec:intro}



As LLM-based agents are increasingly deployed in diverse interactive scenarios, evaluating their effectiveness has become an important research focus. Existing evaluation paradigms have progressively evolved from measuring surface-level response quality to assessing overall task outcomes and interactive behaviors~\cite{xi2025rise,liu2024agentbench,schick2023toolformer}. These advances have substantially improved agent benchmarks and enabled broader real-world deployment~\cite{wang2023voyager,liu2024agentbench}.

However, in interaction-intensive scenarios (e.g., emotional support~\cite{liu2020towards,zhang2026empa}, persuasion~\cite{wang2019persuasion,bozdag2026must}, and conversational recommendation~\cite{liu2020towards,an2025beyond}), agent effectiveness depends not merely on whether a task is ultimately completed, but on how agents shape users' evolving cognitive states across multiple turns~\cite{chawla2023social,qiu2023minddial}. Early evaluation methods primarily rely on response-level metrics, such as BLEU~\cite{papineni2002bleu} and ROUGE~\cite{lin2004rouge}, which mainly capture lexical overlap and fail to evaluate whether responses are semantically appropriate or effectively adapt to users' needs. To overcome this limitations, recent approaches leverage LLMs as evaluators to provide semantic and holistic judgments of conversations. Frameworks such as MT-Bench~\cite{zheng2023judging} and AlpacaEval~\cite{dubois2024length} evaluate overall response quality beyond lexical similarity. However, they typically aggregate multi-turn interactions into final scores, providing limited visibility into the intermediate processes and cognitive dynamics that lead to successful interactions.

To enable process-oriented evaluation beyond static judgments, LLM-based user simulation introduces simulated users for scalable multi-turn interaction evaluation. Early systems~\cite{luo2024duetsim,chang2025chatbench,kong2024platolm} focus on generating realistic user utterances, enabling behavior-level evaluation without costly human participants. More recent approaches incorporate cognitive elements, such as implicit profiles~\cite{wang2025know}, latent representations~\cite{wu2026humanlm}, or belief tracking~\cite{qiu2023minddial}. Nevertheless, existing simulators still lack a unified and interpretable framework for modeling how users' cognitive states evolve throughout interactions.

\begin{table}[h]
\centering

\setlength{\tabcolsep}{4pt}
\renewcommand{\arraystretch}{1.2}
\begin{tabularx}{0.9\linewidth}{| X |c | c | c | c |}
\hline
\textbf{Paradigm}
& \textbf{Sim.}
& \textbf{Cogn.}
& \textbf{Trace}
& \textbf{Process} \\
\hline
Text Metrics
& \ding{55}
& \ding{55}
& \ding{55}
& \ding{55} \\
\hline
LLM-as-Judge
& \ding{55}
& \ding{55}
& \ding{55}
& \ding{55} \\
\hline
User Simulator
& \ding{51}
& $\triangle$
& \ding{55}
& \ding{51} \\
\hline
\textbf{CogWM (Ours)}
& \ding{51}
& \ding{51}
& \ding{51}
& \ding{51} \\
\hline
\end{tabularx}
\vspace{-2mm}
\caption{Comparison of automated evaluation paradigms.}
\label{tab:method_comparison}
\vspace{-3mm}
\end{table}

\begin{figure*}[t!]
\centering
\includegraphics[width=0.95 \textwidth]{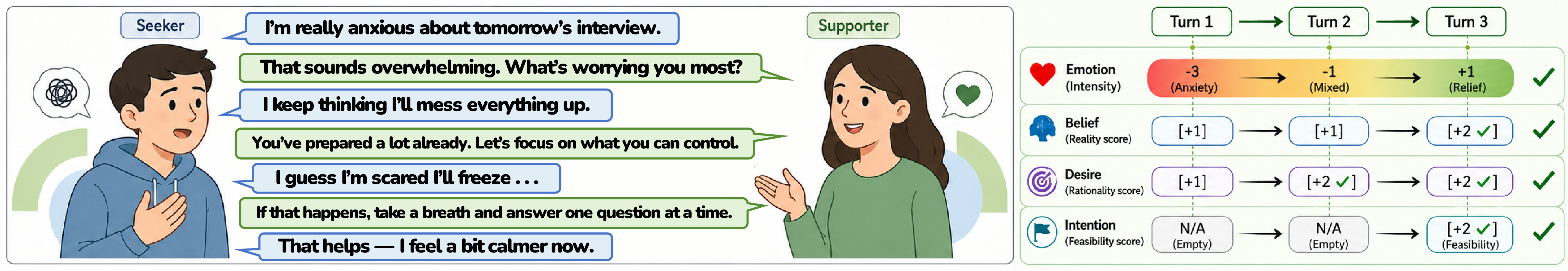}
\vspace{-4mm}
\caption{\textbf{An emotional support dialogue example:} the user transitions from a high-anxiety state (``afraid of messing up'') toward emotional relief through multi-turn interaction, accompanied by observable BDI/E state transitions at each turn.}
\vspace{-5mm}
\label{fig:intro_example}
\end{figure*}


As summarized in Table~\ref{tab:method_comparison}, no existing approach simultaneously supports interactive simulation, explicit multi-dimensional cognitive modeling, traceable state evolution, and process-level evaluation. This gap stems from a fundamental limitation of existing paradigms: they evaluate observable interaction signals without explicitly modeling the latent cognitive dynamics that give rise to user behaviors. Cognitive science suggests that human behaviors are driven by structured mental states rather than isolated responses. The Belief--Desire--Intention (BDI) theory~\cite{bratman1987intention} provides a framework for representing practical reasoning, while the Elaboration Likelihood Model (ELM)~\cite{petty1986elaboration} and Theory of Planned Behavior (TPB)~\cite{ajzen1991theory} further highlight how cognitive changes, emotions, and intentions contribute to behavioral outcomes. These perspectives motivate explicit modeling of users' cognitive trajectories as a foundation for process-level conversational agent evaluation.



To address this, we propose \textbf{Cog}nitive \textbf{W}orld \textbf{M}odel (\textbf{CogWM}), an LLM-based cognitive user model that shifts evaluation from what users say to how their cognitive states evolve. CogWM introduces an explicit Belief, Desire, Intention, and Emotion (BDI/E) representation and jointly predicts cognitive states and utterances at each turn. Crucially, we design the \textbf{Summarize-and-Allocate (SaA)} construction pipeline to provide temporally coherent supervision, which first extracts global cognitive evolution from complete dialogues, then allocates turn-level BDI/E labels aligned with this global trajectory. Unlike independent turn-level annotation, SaA enables CogWM to learn structured cognitive dynamics rather than isolated state-response mappings, ensuring temporal consistency in trajectory prediction.



Leveraging CogWM's structured outputs, we further introduce a progressive BDI/E trajectory evaluation framework that assesses conversational agents across three complementary granularity levels. The turn level constructs task-aware cognitive states at each interaction step, the trajectory level characterizes cognitive evolution from cumulative quality, improvement magnitude, transition consistency, and trajectory stability, and the task level evaluates final behavioral outcomes. This hierarchical design enables interpretable, process-level diagnosis of agent behaviors by revealing how cognitive evolution relates to interaction effectiveness.

Our contributions are summarized as follows.

(1) We propose CogWM, an LLM-based cognitive user model that jointly models user BDI/E states and responses. CogWM achieves superior performance over user simulation baselines in response fidelity and state understanding.

(2) We introduce a progressive BDI/E trajectory evaluation framework that evaluates conversational agents from turn-level cognitive responses, trajectory-level dynamics, and task-level outcomes, enabling traceable analysis.


(3) We evaluate CogWM through free-form interactions with multiple conversational agents, demonstrating that it reveals distinct cognitive trajectories patterns and complements outcome-based evaluation.



\section{Related Work}
\label{sec:related_work}
\subsection{Cognitively Demanding Conversational Tasks}
Recent conversational AI applications increasingly involve cognitively demanding conversational scenarios, including emotional support dialogue~\cite{liu2021towards}, persuasive dialogue~\cite{wang2019persuasion}, and conversational recommendation~\cite{liu2020towards}. Unlike conventional task-oriented dialogue evaluation, which often emphasizes explicit goal completion and task success~\cite{chen2017survey}, these scenarios require agents to understand and influence users' evolving beliefs, desires, intentions, and emotions throughout multi-turn interactions. However, many existing evaluation paradigms mainly assess response quality or final outcomes, providing limited insight into the latent cognitive processes underlying interaction dynamics~\cite{chawla2023social}.

\subsection{Dialogue Evaluation}
Automatic dialogue evaluation has progressed from reference-based metrics to LLM-based assessment and simulation. Traditional metrics such as BLEU~\cite{papineni2002bleu} and ROUGE~\cite{lin2004rouge} measure response similarity but rely heavily on reference answers and cannot reflect user-level cognitive state evolution. LLM-as-Judge approaches, including MT-Bench~\cite{zheng2023judging} and AlpacaEval~\cite{dubois2024length}, improve semantic evaluation but still reduce entire interactions to terminal scores without tracing the underlying influence process.

Recent LLM-based user simulators enable scalable interaction-based evaluation by replacing human participants with simulated users. Methods such as DuetSim~\cite{luo2024duetsim}, ChatBench~\cite{chang2025chatbench}, and PlatoLM~\cite{kong2024platolm} achieve realistic user response generation, while recent studies explore user state modeling through implicit profiles~\cite{wang2025know}, latent representations~\cite{wu2026humanlm}, and belief tracking~\cite{qiu2023minddial}. However, existing approaches mainly model observable behaviors or implicit states, lacking explicit and unified cognitive representations for tracking users' Belief, Desire, Intention, and Emotion.

\subsection{Cognitive Modeling for Interactive Agents}
Cognitive modeling aims to represent latent mental states underlying interactive behaviors. The Belief-Desire-Intention (BDI) framework~\cite{bratman1987intention,georgeff1991modeling} provides a classical model for representing practical reasoning through beliefs, desires, and intentions. Recent studies have explored cognitive state modeling in dialogue systems, including belief-desire representations~\cite{li2021past}, explicit belief-desire graphs~\cite{xu2024beyond}, and sequential BDI prediction in persuasive dialogue~\cite{ma2026think}. Nevertheless, existing methods typically focus on individual cognitive dimensions or implicit state inference, without jointly modeling explicit cognitive states and user responses to capture cognitive evolution during multi-turn conversations.

\section{Cognitive World Model}
\label{sec:framework}

Inspired by cognitive theories that characterize human behaviors as the outcomes of evolving internal mental states~\cite{bratman1987intention}, 
we formulate cognitive user modeling as a latent state transition process, where user behaviors emerge from evolving cognitive states. Given a multi-turn dialogue sequence $(u_1,a_1,\ldots,u_T,a_T)$, CogWM learns the transition:

\begin{equation}
(P,C,H_{<t},S_{t-1}) \rightarrow (S_t,U_t),
\label{eq:cognitive_transition}
\end{equation}

\noindent where $P$ denotes user profile information, $C$ represents task context, $H_{<t}$ is the dialogue history, $S_{t-1}$ and $S_t$ denote previous and current cognitive states, and $U_t$ is the user utterance. This formulation enables CogWM to model how latent cognitive states evolve through interaction and how such evolution is reflected in observable user response.


To represent the latent cognitive state $S_t$, we define a structured state space based on Belief--Desire--Intention (BDI) theory~\cite{bratman1987intention} and extend it with Emotion to capture affective dynamics:

\begin{equation}
S_t=\{B_t,D_t,I_t,E_t\},
\label{eq:bdie_state}
\end{equation}

\noindent where $B_t$, $D_t$, $I_t$, and $E_t$ represent the user's BDI/E states at turn $t$, respectively. This representation provides an interpretable state space for tracking users' cognitive evolution across multi-turn interactions. Table~\ref{tab:dbi_role} summarizes the definitions and roles of BDI/E dimensions.

\newcolumntype{J}[1]{>{\arraybackslash\hsize=#1\hsize}X}

\begin{table}[t!]
\centering

\small
\renewcommand{\arraystretch}{1.20}
\setlength{\tabcolsep}{4pt}
\begin{tabularx}{\linewidth}{
>{\centering\arraybackslash}m{0.35cm}
J{0.9}
J{1.1}
}
\toprule
&
\multicolumn{1}{c}{\textbf{Definition}} &
\multicolumn{1}{c}{\textbf{Role}} \\
\midrule
\rotlabel[-6mm]{Belief} &
What the user believes, assumes or interprets. Beliefs may be correct, incorrect, or unverifiable. &
Tracks how the agent reshapes users' understanding or corrects misconceptions. \\
\midrule
\rotlabel[-3mm]{Desire} &
What the user wants, needs, prefers, or avoids. &
Tracks changes in users' needs, preferences, and motivations. \\
\midrule
\rotlabel[-3mm]{Intention} &
What the user plans, decides, accepts, rejects, or commits to doing. &
Tracks decision formation and behavioral commitment. \\
\midrule
\rotlabel[-4mm]{Emotion} &
The user's current affective state and its intensity. &
Tracks affective responses and emotional regulation during interaction.  \\
\bottomrule
\end{tabularx}
\vspace{-2mm}
\caption{Definitions of BDI/E Cognitive Dimensions.}
\label{tab:dbi_role}
\vspace{-2mm}
\end{table}

For trajectory analysis, each cognitive dimension is represented as $X_t=(z_t,q_t)$, where $z_t$ denotes the semantic state description and $q_t$ is a task-aware score measuring alignment with scenario-specific objectives. Table~\ref{tab:task_bdi_scores} provides task-specific score definitions in three representative scenarios.

\begin{figure}[t!]
\centering
\includegraphics[width=0.9\columnwidth]{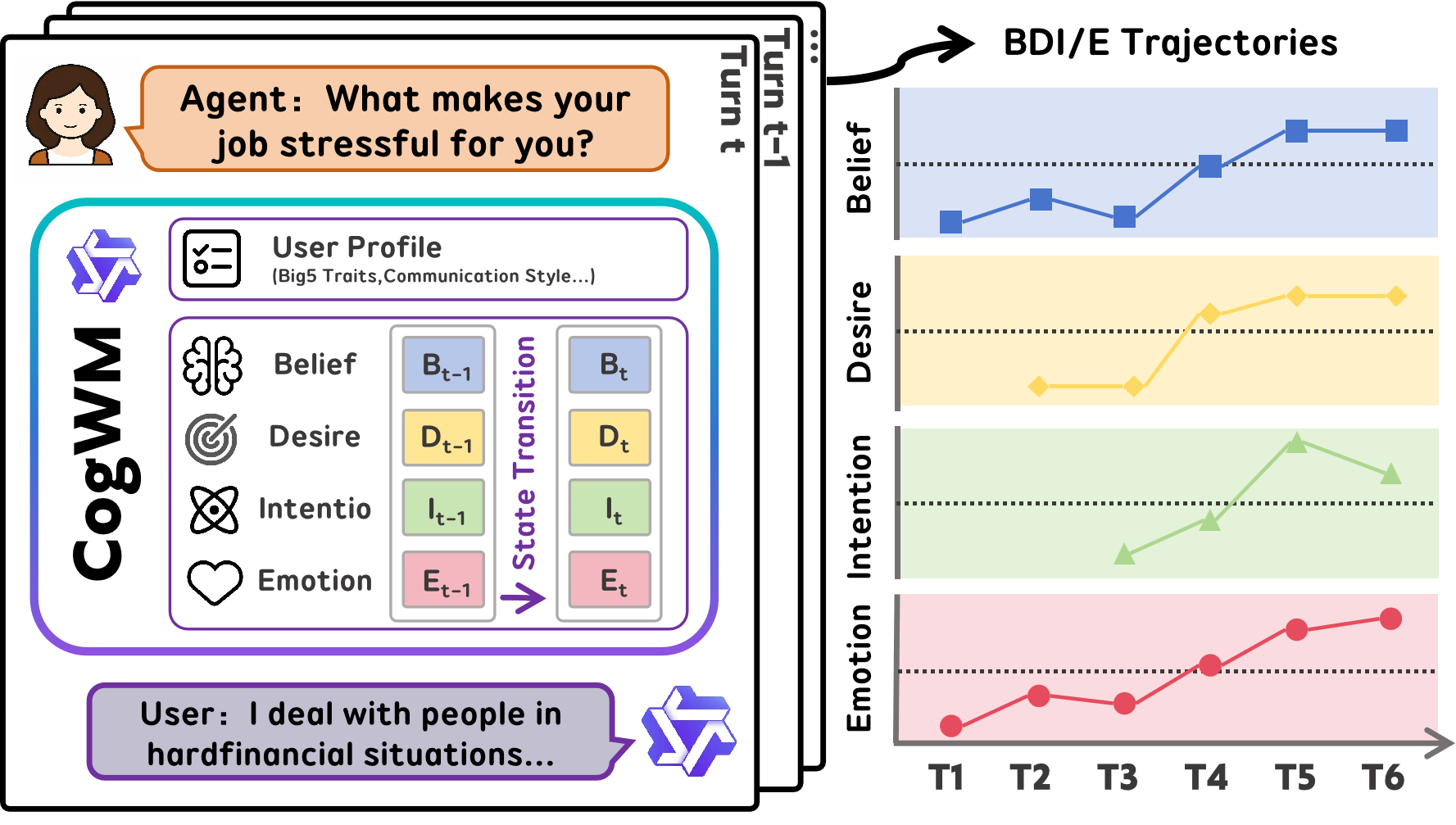}
\vspace{-2mm}
\caption{\textbf{Overview of CogWM.} Given user profiles, task context, and dialogue history, CogWM predicts evolving BDI/E states and corresponding user responses, forming explicit cognitive trajectories.}
\vspace{-4mm}
\label{fig:overview}
\end{figure}

Based on the above cognitive state formulation, CogWM jointly predicts the current BDI/E state and the corresponding user utterance. As shown in Figure~\ref{fig:overview}, CogWM models cognitive states as intermediate representations between conversational context and observable user behaviors, generating structured BDI/E trajectories throughout interactions.

\begin{table}[t!]
\centering
\small
\begin{tabularx}{0.95\columnwidth}{X c c c}
\toprule
\textbf{Scenario} & \textbf{Belief} & \textbf{Desire} & \textbf{Intention} \\
\midrule
Support & reality & rationality & feasibility \\
Persuasion & impact & prosocial & commitment \\
Recommendation & alignment & satisfaction & acceptance \\
\bottomrule
\end{tabularx}
\vspace{-2mm}
\caption{Task-specific BDI Quantification Scores}
\label{tab:task_bdi_scores}
\end{table}


\vspace{-2mm}

\section{Cognitive Supervision Construction}
\label{sec:data}

Existing dialogue datasets provide rich multi-turn interactions but lack explicit annotations of users' latent cognitive evolution. To enable CogWM to learn cognitive trajectories rather than isolated state-response mappings, we construct trajectory-aware supervision from multi-turn dialogue data. We collect conversations from four public dialogue datasets,  including DailyDialog~\cite{li2017dailydialog} for open-domain conversations and three cognitively demanding scenarios: emotional support (ESConv)~\cite{liu2021towards}, charitable donation persuasion (P4G)~\cite{wang2019persuasion}, and conversational recommendation (DuRecDial)~\cite{liu2020towards}. 


User cognitive states evolve across dialogue turns rather than appearing as independent responses. To construct structured BDI/E supervision, we propose Summarize-and-Allocate (SaA), a two-stage annotation pipeline. First, the global summarization stage extracts dialogue-level BDI/E states and corresponding task-specific scores from the complete conversation. Second, the temporal allocation stage assigns these states to individual turns with evidence grounding, preserving the temporal evolution of users' cognitive states. Furthermore, explicit \texttt{bdi\_link} constraints encode dependencies among BDI, allowing CogWM to learn structured cognitive transitions from data. Human evaluation verifies the annotation quality, achieving BDI recall of 0.79, precision of 0.77, and emotion Cohen's $\kappa=0.69$.


To provide personalized conditioning signals, we annotate each dialogue with user profiles following prior studies on profile inference from dialogue~\cite{wang2025know}. The profiles include personality traits, communication style, decision-making style, and task-related characteristics. 


Finally, we construct 150,454 user-turn training samples. CogWM is jointly optimized for cognitive state prediction and response generation using the objective $p(U_t,S_t|\mathcal{D}_{<t},P,C)$. We initialize CogWM with Qwen3-14B-Instruct~\cite{yang2025qwen3} and apply LoRA fine-tuning~\cite{hu2022lora}. Details of dataset and training are provided in the Supplementary Material.

\section{Cognitive Trajectory Evaluation}
\label{sec:evaluation}
CogWM provides explicit BDI/E trajectories of users throughout multi-turn interactions, enabling evaluation beyond isolated responses and terminal outcomes. Based on these structured cognitive trajectories, we develop a progressive evaluation framework with three complementary levels: (1) turn-level cognitive quantification measures the instantaneous cognitive state at each interaction step (\S\ref{sec:evaluation:turn}); (2) trajectory-level dynamics characterize the temporal evolution of cognitive states across conversations (\S\ref{sec:evaluation:trajectory}); and (3) task-level outcomes evaluate whether the interaction achieves the intended objective (\S\ref{sec:evaluation:task}). 

\subsection{Turn-level Cognitive Quantification.}
\label{sec:evaluation:turn}
At each turn, the four task-aware BDI/E scores are aggregated into a composite cognitive score:

\vspace{-2mm}
\begin{equation}
\mathrm{Comp}_t=w_EE_t+w_BB_t+w_DD_t+w_II_t ,
\end{equation}

\noindent where  $B_t$, $D_t$, $I_t$, and $E_t$ denote the quantified BDI/E states, respectively. The weights reflect task-specific objectives: emotional support assigns greater weight to Emotion, whereas persuasion and recommendation emphasize the three BDI dimensions. 

\subsection{Trajectory-level Cognitive Evolution} 
\label{sec:evaluation:trajectory}
$\{\mathrm{Comp}_t\}_{t=1}^{T}$ forms the scalar cognitive trajectory for subsequent analysis. We characterize each trajectory using five complementary metrics. AUC measures the cumulative cognitive quality over the entire trajectory through numerical integration. PSR quantifies the frequency of positive transitions between consecutive turns. TS-$\Delta$ estimates the overall improvement magnitude based on a robust Theil--Sen slope, reducing sensitivity to local fluctuations and endpoint noise. $\Delta$ captures the net change between substantive initial and final states, while Vol measures trajectory stability by quantifying the variance of turn-level state transitions. The complete formulations are provided in the Supplementary Material.


Based on 3,600 interaction trajectories, we analyze the pairwise correlations among these metrics and select three complementary metrics with low redundancy, AUC, PSR, and TS-$\Delta$, for CTS construction. All pairwise correlations among the selected metrics satisfy $|r|\leq0.24$. The Composite Trajectory Score (CTS) is defined as:

\vspace{-2mm}
\begin{equation}
\mathrm{CTS}
=
0.40\cdot\mathrm{AUC}
+
0.30\cdot\mathrm{PSR}
+
0.30\cdot\mathrm{TS}\text{-}\Delta .
\end{equation}

\noindent Here, AUC captures cumulative cognitive quality, PSR reflects the frequency of positive state transitions, and TS-$\Delta$ measures the robust overall improvement magnitude. Together, these three metrics provide complementary signals for cognitive trajectory evolution. Weight sensitivity analysis further shows that the agent ranking remains highly stable across alternative configurations (Kendall's $W=0.954$).

\subsection{Task-level Behavioral Outcome.} 
\label{sec:evaluation:task}
While CTS measures cognitive influence reflected in user state trajectories, behavioral outcomes provide another perspective. Therefore, we introduce GoodOutcome (GO), an LLM-based binary judgment that evaluates whether the final dialogue outcome satisfies the task objective. The criteria are scenario-dependent: emotional support considers emotional improvement or actionable coping plans, persuasion considers donation decisions, and conversational recommendation considers recommendation acceptance.

\vspace{-1mm}
\section{Experiments}
\label{sec:exp}
We evaluate CogWM from two complementary perspectives: (1) We assess CogWM as a user simulator through three experiments: Joint State-Utterance Generation evaluates its ability to jointly model user utterances and BDI/E states; State Understanding examines its capability to infer latent user cognitive states from dialogue contexts; and Context Conditioning Ablation analyzes the contribution of different contextual signals (\S\ref{sec:exp:fidelity}, \S\ref{sec:exp:e0}, and \S\ref{sec:exp:ablation}); (2) We demonstrate its utility as an evaluation framework by analyzing conversational agents through multi-turn interactions (\S\ref{sec:exp:agent_discrimination}). Additional statistical analyses, robustness studies, and case studies are provided in the supplementary material.

\subsection{Experimental Setup}
\label{sec:exp:setting}
\noindent\textbf{Joint State-Utterance Generation.}
All simulation experiments use CogWM-14B and run on a single A100-80G GPU for inference. The test set comprises 7,563 user-turn samples, including 3,765 from DailyDialog, 1,137 from ESConv, 508 from P4G, and 2,153 from DuRecDial.

\begin{table*}[t!]
\centering
\small

\setlength{\tabcolsep}{1.5pt}
\begin{tabular}{l
c c c c c
c c c
c c c c c c
}
\toprule
\multirow{2}{*}{\textbf{Model}}
& \multicolumn{5}{c}{\textbf{State Prediction}}
& \multicolumn{3}{c}{\textbf{Utterance}}
& \multicolumn{6}{c}{\textbf{LLM Judge}} \\
\cmidrule(lr){2-6} \cmidrule(lr){7-9} \cmidrule(lr){10-15}
& E-A↑ & B-S↑ & D-S↑ & I-S↑ & E-M↓
& R-L↑ & B-4↑ & LenR
& Emo↑ & Bel↑ & Des↑ & Int↑ & Utt↑ & Coh↑ \\
\midrule
GPT-5.5 & 36.4 & .474 & .433 & .442 & 0.46 & .237 & .043 & 1.82 & 2.34 & 1.90 & 2.09 & 1.74 & 2.62 & \textbf{4.71} \\
DS-V4-Pro & 31.0 & .462 & .445 & .412 & 0.52 & .288 & .060 & 1.81 & 2.13 & 1.87 & 2.21 & 1.93 & 2.73 & 4.57 \\
Dual-LLM & 30.9 & .500 & .446 & .445 & 0.47 & .253 & .054 & 2.12 & 2.14 & 1.84 & 2.27 & 1.95 & 2.67 & 4.49 \\
\midrule
Qwen3-14B & 28.1 & .384 & .358 & .416 & 0.63 & .196 & .033 & 2.26 & 2.37 & 1.79 & 2.06 & 1.82 & 2.41 & 4.45 \\
utt-only-14B & 12.2 & .285 & .340 & .264 & — & \textbf{.427} & \textbf{.159} & 1.03 & 2.22 & 1.24 & 1.17 & 1.31 & 1.13 & 2.43 \\
\textbf{CogWM-14B} & \textbf{77.6} & \textbf{.773} & \textbf{.705} & \textbf{.458} & \textbf{0.29} & .417 & .150 & \textbf{1.01} & \textbf{4.49} & \textbf{4.16} & \textbf{4.33} & \textbf{4.75} & \textbf{3.32} & 4.50 \\
\bottomrule
\end{tabular}%
\vspace{-2mm}
\caption{Joint state-utterance generation results of CogWM.}
\label{tab:main}
\vspace{-4mm}
\end{table*}

We evaluate simulation fidelity from three aspects: BDI/E state prediction, response generation quality, and state--utterance consistency. E-Acc\textbf{(E-A)} measures the accuracy of predicted emotion labels; MAE\textbf{(-M)} measures the prediction error of BDI/E scores; Sim\textbf{(-S)} evaluates the semantic similarity between predicted and reference BDI/E states using all-MiniLM-L6-v2 as the embedding model. ROUGE-L\textbf{(R-L)}, BLEU-N\textbf{(B-n)}, and LengthRatio\textbf{(LenR)} ($\frac{|pred|}{|gt|}$, $\approx 1$ optimal) measure lexical similarity and length consistency of generated responses. We further employ Qwen3.7-Plus as an LLM judge to evaluate six dimensions on a 1--5 scale: Emo, Bel, Des, Int, Utt, and Coh. The first five dimensions measure alignment with reference emotions, cognitive states, and utterances, while Coh evaluates the consistency between predicted cognitive states and generated responses.

Due to the sparsity of BDI/E annotations (61.7\% of belief, 68.1\% of desire, and 91.4\% of intention annotations are empty), all BDI/E-related metrics are computed only on turns with non-empty ground-truth states, resulting in 3,342 evaluated turns. 

\noindent\textbf{Simulation Baselines.}
We compare CogWM with five types of simulator baselines. \textbf{Qwen3-14B}~\cite{yang2025qwen3} is evaluated under zero-shot prompting without fine-tuning, measuring the intrinsic BDI/E modeling capability of the base model. \textbf{utt-only-14B} uses the same base model and training configuration as CogWM-14B but is trained only for user utterance generation, serving as an utterance-generation upper bound. \textbf{GPT-5.5} and \textbf{DeepSeek-V4-Pro}~\cite{xu2026deepseek} are included as strong API-based baselines. \textbf{Dual-LLM} adapts the dual-LLM verification paradigm from DuetSim~\cite{luo2024duetsim} with task-specific prompting and is implemented using GPT-5.5. All simulator baselines follow the same output schema as CogWM.

\noindent\textbf{Conversational Agent Evaluation.}
We apply CogWM to measure how different conversational agents influence users' cognitive trajectories. Following prior conversational dialogue studies~\cite{li2017dailydialog,liu2021towards,liu2020towards}, we construct prompts to instantiate six representative LLM-based agents, including GPT-5.5, Llama-4-Scout, DeepSeek-V4-Pro, Claude-Opus-4.7, Gemini-3-Pro, and Qwen3.7-Plus, while CogWM-14B serves as the user simulator.

We conduct $6$ agents $\times$ $2$ prompting conditions (with and without Chain-of-Thought, CoT) $\times$ $3$ tasks $\times$ $100$ dialogues, resulting in 3,600 multi-turn interactions. The first two turns use real dialogue contexts for cold start. From the eighth turn onward, an LLM-based intent detection module determines whether the dialogue has naturally reached completion, with a maximum dialogue length of 20 turns.

Agents are evaluated using the three-tier framework described in \S\ref{sec:evaluation}. The primary metric is \textbf{CTS}, while GoodOutcome \textbf{(GO)} evaluates whether the dialogue achieves the task-specific behavioral objective.

\vspace{-2mm}
\subsection{Joint State-Utterance Generation}
\label{sec:exp:fidelity}

Table~\ref{tab:main} presents the simulation fidelity results. Overall, CogWM achieves substantially stronger cognitive state understanding performance than existing baselines while maintaining competitive user response generation quality, demonstrating that explicit cognitive supervision enables more accurate state modeling beyond response-level imitation.

\textbf{BDI/E State Prediction.}
CogWM achieves the strongest performance across all cognitive state prediction metrics. It obtains 77.6\% emotion accuracy, substantially outperforming all baselines. For semantic alignment of cognitive states, CogWM achieves B-Sim, D-Sim, and I-Sim scores of 0.773, 0.705, and 0.458, respectively, consistently exceeding both API-based and prompting-based baselines. The improvements over baselines are statistically significant under bootstrap testing with 10,000 resamples, with large effect sizes observed on major metrics (e.g., Cohen's $d=1.25$ for B-Sim and $d=0.91$ for E-Acc). These results indicate that explicit BDI/E supervision enables CogWM to capture latent cognitive states that cannot be reliably inferred through zero-shot prompting alone.

\textbf{Response Generation Quality.}
Despite the additional objective of cognitive state modeling, CogWM maintains strong response generation capability. It achieves ROUGE-L of 0.417 and BLEU-4 of 0.150, outperforming all API-based baselines. More importantly, CogWM produces responses with better length consistency, achieving a LengthRatio of 1.01 compared with GPT-5.5 (1.82) and Qwen3-14B (2.26). This indicates that CogWM generates responses closer to the reference user response distribution, while avoiding the excessive verbosity observed in some baselines.

\textbf{Effect of Joint State-Utterance Training.}
To examine whether cognitive state modeling can emerge from response generation objectives alone, we compare CogWM with utt-only-14B, which uses the same base model, training data, and optimization configuration but is trained only for user utterance generation. Although utt-only-14B achieves slightly better surface-level response metrics (ROUGE-L: 0.427 vs. 0.417; BLEU-4: 0.159 vs. 0.150), it fails to recover meaningful cognitive states, obtaining only 12.2\% emotion accuracy and substantially lower BDI similarity scores. In contrast, CogWM sacrifices less than 6\% relative response generation performance while achieving large improvements in cognitive state prediction. This demonstrates that user cognitive states cannot be implicitly acquired from language generation objectives and require explicit cognitive supervision.

\textbf{Cognitive Grounding Evaluation.}
LLM-based semantic evaluation further confirms the quality of CogWM's structured outputs. CogWM achieves the highest scores on emotion, belief, desire, and intention alignment, with scores of 4.49, 4.16, 4.33, and 4.75, respectively. API-based models obtain relatively higher coherence scores, but their advantages do not translate into accurate cognitive state prediction. This indicates that linguistic consistency and cognitive grounding represent different capabilities: a model may generate internally coherent responses while failing to accurately model the underlying user state. Such distinction motivates the use of explicit cognitive state tracking for downstream social influence evaluation.

\vspace{-2mm}
\subsection{Cognitive State Understanding} \
\vspace{-4mm}
\label{sec:exp:e0}

To isolate cognitive state inference from response generation, we conduct an oracle comprehension experiment, where models are provided with dialogue context and ground-truth user utterances to predict BDI/E states. This setting evaluates whether models can recover latent user cognition when the surface response is explicitly available.

\begin{table}[ht]
\centering

\small
\setlength{\tabcolsep}{1.5pt}
\begin{tabularx}{0.95\columnwidth}{X c c c c c c c c}
\toprule
\textbf{Model} & \textbf{E-A} & \textbf{B-Sim} & \textbf{D-Sim} & \textbf{I-Sim}
& \textbf{L-E} & \textbf{L-B} & \textbf{L-D} & \textbf{L-I} \\
\midrule
GPT-5.5
& 42.3 & 0.489 & 0.457 & \textbf{0.471}
& 2.59 & 1.97 & 2.29 & 2.75 \\

DSv4-Pro
& 31.9 & 0.468 & 0.430 & 0.425
& 2.38 & 1.80 & 2.28 & 2.54 \\

\midrule

Qwen3-14B
& 32.4 & 0.398 & 0.361 & 0.419
& 2.02 & 1.54 & 1.87 & 2.02 \\

\textbf{CogWM-14B}
& \textbf{74.5} & \textbf{0.768} & \textbf{0.675} & 0.452
& \textbf{4.46} & \textbf{3.67} & \textbf{3.37} & \textbf{3.42} \\

\bottomrule
\end{tabularx}
\vspace{-2mm}
\caption{State Understanding Results}
\label{tab:e0}
\vspace{-3mm}
\end{table}

Table~\ref{tab:e0} presents the results, CogWM achieves the best performance across almost all cognitive state metrics, obtaining 74.5\% emotion accuracy, 0.768 B-Sim, and 0.675 D-Sim, substantially outperforming both API-based and open-source baselines. The LLM-based evaluation further confirms this advantage, where CogWM achieves higher alignment scores on emotion, belief, desire, and intention dimensions. These results demonstrate that explicit BDI/E supervision enables CogWM to recover latent user cognition beyond surface-level linguistic understanding.

Compared with the joint generation setting in Table~\ref{tab:main}, CogWM maintains consistent cognitive understanding performance when user utterances are directly provided (E-A: 77.6\% vs. 74.5\%). In contrast, zero-shot LLMs benefit from oracle utterances but remain substantially behind CogWM, suggesting that their limitation mainly lies in the absence of explicit cognitive state modeling rather than insufficient access to user responses.

Intention remains the most challenging dimension across models due to its future-oriented nature and sparse annotations. Nevertheless, CogWM achieves competitive intention alignment (I-Sim: 0.452) and the highest LLM-based intention score (L-I: 3.42), suggests that structured BDI/E supervision facilitates the modeling of latent decision states beyond directly observable utterances.

\subsection{Context Conditioning Ablation} \
\label{sec:exp:ablation}
We investigate the contribution of different conditioning signals through input-level ablation, where each component is removed at inference time while keeping model parameters unchanged. Table~\ref{tab:ablation} reports six core metrics. The impact of each removal is further quantified using paired Cohen's $d$ computed from per-sample differences, where $|d|\geq0.20$ with a 95\% CI excluding zero indicates a measurable effect.

\begin{table}[ht]
\centering
\small
\setlength{\tabcolsep}{2pt}
\begin{tabularx}{0.82\columnwidth}{X c c c c c c}
\toprule
\textbf{Model}
& \textbf{E-A}
& \textbf{B-Sim}
& \textbf{D-Sim}
& \textbf{I-Sim}
& \textbf{R-L}
& \textbf{B-4} \\
\midrule
-- Task
& 74.2 & .770 & .686 & .538
& .377 & .148 \\
-- Big5
& 73.5 & .761 & .683 & .539
& .376 & .146 \\
-- Comm/Dec
& 73.6 & .757 & .675 & .533
& .374 & .140 \\
-- Profile
& 72.5 & .748 & .672 & .506
& .365 & .140 \\
\hline
-- All
& 60.7 & .712 & .591 & .399
& .210 & .040 \\
\textbf{Full}
& \textbf{77.6}
& \textbf{.773}
& \textbf{.705}
& \textbf{.554}
& \textbf{.417}
& \textbf{.150} \\
\bottomrule
\end{tabularx}
\vspace{-2mm}
\caption{Context conditioning ablation results.}
\label{tab:ablation}
\vspace{-2mm}
\end{table}

\textbf{Individual context signals provide complementary information.}
Removing individual conditioning components, including task context, Big Five personality, or communication/decision style, results in only minor performance changes ($|d|\leq0.10$). This indicates that CogWM does not rely on any single contextual source, as user characteristics and situational information can be partially compensated by dialogue history and other available signals. Removing the complete user profile ($-$Profile) causes a larger but still moderate degradation, reducing E-Acc from 77.6\% to 72.5\% and B-Sim from 0.773 to 0.748, suggesting that explicit user profiles provide additional user-specific grounding.

\textbf{Removing all contextual information leads to substantial degradation.}
When all profile and task context signals are removed ($-$All), performance decreases substantially across both cognitive and language metrics. In particular, ROUGE-L drops from 0.417 to 0.210 ($d=0.63$), and BLEU-4 decreases from 0.150 to 0.040 ($d=0.40$), demonstrating the importance of contextual grounding for personalized response generation. Cognitive state prediction also degrades significantly, including B-Sim ($d=0.29$), D-Sim ($d=0.36$), and E-Acc ($d=0.23$), showing that explicit user and task context provides information beyond dialogue history alone. Overall, these results validate that heterogeneous contextual signals act as complementary grounding sources for CogWM's cognitive modeling and user simulation.

\begin{table*}[t!]
\centering

\small
\setlength{\tabcolsep}{1.5pt}
\begin{tabularx}{0.895\textwidth}{c | c | c | c | c c c c c | c c c c | c | c c c c}
\toprule
\textbf{\#} & \textbf{Model} & \textbf{Setting}
& \textbf{CTS} & \textbf{AUC} & TS-$\Delta$ & PSR & $\Delta$ & Vol
& E-$\Delta$ & B-$\Delta$ & D-$\Delta$ & I-$\Delta$
& nz\% & GO & GO-E & GO-P & GO-D \\
\midrule

1 & Llama-4-Scout & Vanilla
& +.075 & \textbf{-.006} & +.071 & .354 & +.430 & .577
& +.637 & +.158 & +.277 & +.270
& 37.1 & 68.7 & 79 & 66 & 61 \\

2 & Llama-4-Scout & CoT
& \textbf{+.083} & -.012 & +.078 & .367 & +.441 & .559
& \textbf{+.780} & +.216 & +.322 & +.238
& \textbf{38.2} & 72.7 & 84 & 66 & 68 \\

3 & Qwen3.7-Plus & Vanilla
& +.073 & -.024 & \textbf{+.096} & \textbf{.358} & +.407 & .530
& +.198 & +.191 & \textbf{+.409} & \textbf{+.353}
& 38.0 & 78.0 & 80 & 79 & 75 \\

4 & Qwen3.7-Plus & CoT
& +.039 & -.023 & +.005 & .346 & \textbf{+.472} & .563
& +.493 & +.219 & +.402 & +.261
& 36.2 & \textbf{79.3} & \textbf{90} & 68 & \textbf{80} \\

5 & Claude-Opus-4.7 & Vanilla
& +.031 & -.030 & +.009 & .323 & +.382 & .537
& +.248 & +.195 & +.398 & +.177
& 34.8 & 66.0 & 70 & 58 & 70 \\

6 & Claude-Opus-4.7 & CoT
& +.069 & -.010 & +.069 & .352 & +.391 & .565
& +.310 & +.161 & +.305 & +.250
& 36.4 & 73.3 & 87 & 67 & 66 \\

7 & DeepSeek-V4-Pro & Vanilla
& +.039 & -.035 & +.049 & .325 & +.382 & .519
& +.373 & \textbf{+.228} & +.293 & +.127
& 35.6 & 72.0 & 84 & 61 & 71 \\

8 & DeepSeek-V4-Pro & CoT
& +.054 & -.026 & +.071 & .336 & +.364 & .526
& +.272 & +.182 & +.394 & +.233
& 35.3 & 70.7 & 83 & 59 & 70 \\

9 & GPT-5.5 & Vanilla
& +.036 & -.035 & +.029 & .316 & +.349 & .500
& +.365 & +.130 & +.241 & +.173
& 35.4 & 68.7 & 80 & 68 & 58 \\

10 & GPT-5.5 & CoT
& +.066 & -.022 & +.113 & .322 & +.405 & .523
& +.380 & +.197 & +.332 & +.190
& 35.1 & 74.7 & 84 & \textbf{82} & 58 \\

11 & Gemini-3-Pro & Vanilla
& +.014 & -.065 & +.015 & .276 & +.278 & .430
& +.618 & +.104 & +.078 & +.153
& 30.4 & 44.7 & 54 & 19 & 61 \\

12 & Gemini-3-Pro & CoT
& +.026 & -.045 & +.039 & .278 & +.292 & .475
& +.453 & +.110 & +.194 & +.113
& 30.7 & 47.3 & 70 & 20 & 52 \\
\bottomrule
\end{tabularx}
\vspace{-2mm}
\caption{Conversational agent comparison based on cognitive trajectory evaluation.}
\label{tab:cts_full}
\vspace{-4mm}
\end{table*}

\subsection{Conversational Agent Evaluation}
\label{sec:exp:agent_discrimination}


Using CogWM as a cognitive evaluator, we analyze how users' cognitive states evolve under interactions with different conversational agents through multi-turn interactions. Table~\ref{tab:cts_full} presents the results. We further perform paired bootstrap analysis with 10,000 resamples to evaluate ranking stability and statistical significance.

\textbf{CTS Ranking and Cognitive Trajectory Patterns.}
Bootstrap analysis reveals a stable three-tier performance structure. Llama-4-Scout and Qwen3.7-Plus achieve the highest CTS scores and are statistically tied ($p=0.425$), while both significantly outperform DeepSeek-V4-Pro ($p=0.024$--$0.036$). DeepSeek-V4-Pro, GPT-5.5, and Claude-Opus-4.7 form the middle tier, with no significant differences among them ($p=0.32$--$0.66$). In contrast, Gemini-3-Pro consistently forms the lowest tier, achieving significantly lower CTS than all other agents ($p<0.005$). Ranking stability analysis is consistent with this grouping: Qwen and Llama obtain the highest probabilities of ranking first ($P(\#1)=55.1\%$ and $43.4\%$), respectively, while Gemini remains ranked last in most bootstrap samples ($P(\#6)=92.4\%$).

Beyond overall ranking, CTS reveals distinct cognitive trajectory patterns among agents. Llama-4-Scout demonstrates balanced improvements across cumulative quality, transition consistency, and endpoint gains, corresponding to an \textbf{All-round} pattern. Qwen3.7-Plus achieves the largest endpoint improvement ($\Delta=+0.472$ under CoT) but relatively lower cumulative trajectory quality, representing a \textbf{Sprint} pattern. Claude-Opus-4.7 exhibits stable but moderate cognitive evolution, corresponding to a \textbf{Steady} pattern, whereas Gemini-3-Pro shows limited cognitive state evolution throughout interactions, forming a \textbf{Hollow} pattern. These patterns demonstrate that trajectory-based evaluation captures qualitative differences beyond overall agent ranking.

\textbf{CTS and GO Capture Complementary Aspects of Interaction Effectiveness.}
CTS characterizes cognitive evolution during multi-turn interactions, whereas GO reflects final task achievement. Their divergence indicates that process-level cognitive changes and final outcomes capture different aspects of conversational effectiveness. Specifically, Llama-4-Scout achieves the highest CTS but only moderate GO performance (68.7\%), while Qwen3.7-Plus obtains the highest GO score (78.0\%) despite ranking second in CTS. These results suggest that cognitive improvement and behavioral success are related but not equivalent capabilities. Therefore, outcome-only evaluation may overlook meaningful differences in how agents perform throughout interaction.

\textbf{Task-specific Cognitive Evolution Patterns.}
CogWM enables analysis of which cognitive dimensions are associated with successful interactions across different tasks. While CTS summarizes overall cognitive evolution, BDI/E decomposition reveals task-dependent patterns. In emotional support, stronger agents primarily exhibit improvements in emotional regulation and coping-related intentions. In persuasion, effective interactions are associated with changes in beliefs, desires, and intentions corresponding to stronger behavioral commitment signals. In recommendation scenarios, preference alignment and decision consistency become dominant cognitive dimensions. These findings indicate that cognitive evolution follows different patterns across conversational tasks rather than a universal trajectory.

\textbf{CoT Effects on Cognitive Trajectories.}
Chain-of-Thought prompting produces heterogeneous effects across agents. GPT-5.5 benefits most from CoT, showing notable improvements in CTS and trajectory trends, suggesting that additional reasoning steps may benefit cognitive state estimation for some models. In contrast, stronger agents such as Llama-4-Scout and Qwen3.7-Plus obtain limited or inconsistent gains, indicating that explicit reasoning does not uniformly improve cognitive trajectory modeling. The effect of CoT is also task-dependent, with larger improvements observed in emotional support scenarios and limited changes in recommendation tasks.

\textbf{Evaluation Insights from CTS.}
BDI/E trajectory decomposition reveals evaluation signals beyond terminal outcomes. We observe a consistent divergence between cognitive and emotional dimensions: in DuRecDial, agents improve composite cognitive scores (Comp-$\Delta$) while decreasing emotional scores (E-$\Delta$), reflecting simultaneous preference refinement and emotional tension. Similarly, persuasive scenarios exhibit measurable emotional trade-offs, where most agents achieve positive outcomes while being associated with negative emotional trajectories. These findings demonstrate that conversational effectiveness cannot be fully characterized by a single outcome score and motivate dimension-wise trajectory tracking.

This divergence further justifies the task-specific weighting design of Comp$_t$. Without task-aware weighting, cognitive gains in DuRecDial would be obscured by emotional declines, while emotion-driven improvements in ESConv would be diluted by relatively flat cognitive dimensions. By assigning $w_E=0.50$ in ESConv and $w_E=0.10$ in other tasks, the framework emphasizes the most informative dimensions for each conversational scenario.

\section{Limitations and Future Work} 
\noindent\textbf{Annotation quality.}
Although SaA provides reliable cognitive supervision, capturing subtle cognitive states remains challenging. Future work will explore more scalable validation and annotation strategies to improve data quality.

\noindent\textbf{Human-grounded evaluation.}
The correspondence between inferred BDI/E trajectories and human perceptions requires further validation. Future work will investigate whether cognitive trajectory patterns and agent typologies correspond to users' subjective experiences.

\noindent\textbf{From evaluation to training.}
Future work will explore extending CogWM toward interactive training, where BDI/E trajectories may provide dense and interpretable signals for optimizing conversational agents.

\section{Conclusion}
\label{sec:conclusion}

This paper presents CogWM, an LLM-based cognitive user model for evaluating conversational agents through explicit BDI/E cognitive state trajectories. We further establish a progressive three-level evaluation framework that analyzes agents from turn-level cognitive responses, trajectory-level cognitive evolution, and task-level outcomes. Interactions with six state-of-the-art LLMs demonstrate that CogWM reveals distinct cognitive trajectories patterns and complements outcome-based evaluation.



\bibliography{aaai2027}

\newpage

\appendix

\section*{Supplementary Materials}

This supplementary material provides additional analyses, experimental details,
and qualitative examples to further validate CogWM.
An overview of the included materials is provided in Table~\ref{tab:supp_overview}.

\begin{table}[h]
\centering
\small
\caption{Overview of supplementary materials.}
\label{tab:supp_overview}
\begin{tabularx}{0.9\linewidth}{ X X }
\toprule
\textbf{Section} & \textbf{Content} \\
\midrule
\textbf{Broader Impact}
(\S\ref{app:broader_impact})
&
Discusses the potential societal impact, ethical considerations, 
and responsible use of CogWM.
\\
\rowcolor{gray!10}
\textbf{Dataset and Annotation Analysis}
(\S\ref{app:data})
&
Provides detailed statistics and quality analysis of the SaA Dataset.
\\

\textbf{Cognitive Trajectory Evaluation}
(\S\ref{app:cts_detail})
&
Provides details of evaluation metrics.
\\
\rowcolor{gray!10}
\textbf{Metric Sensitivity Study}
(\S\ref{app:sensitivity})
&
Analyzes the sensitivity and robustness of the proposed evaluation metrics.
\\

\textbf{Training Settings}
(\S\ref{app:training})
&
Provides training hyperparameters of CogWM.
\\
\rowcolor{gray!10}
\textbf{Extended Experimental Results}
(\S\ref{app:exp})
&
Reports additional quantitative results.
\\

\textbf{Case Studies}
(\S\ref{app:cases})
&
Presents qualitative examples illustrating cognitive state tracking
and failure modes of existing approaches.
\\
\rowcolor{gray!10}
\textbf{Supplementary Figures}
(\S\ref{app:figures})
&
Provides additional visualization results and trajectory analyses.
\\

\bottomrule
\end{tabularx}
\end{table}

\section{Broader Impact}
\label{app:broader_impact}

\subsection{Generalizability of Cognitive Modeling} 
The BDI/E framework and SaA annotation method can be generalized to any dialogue scenario where understanding why a person speaks or acts a certain way is needed. This includes tracking patients' health beliefs and help-seeking intentions in medical consultations, students' knowledge beliefs and learning motivation in educational tutoring, and negotiators' bottom lines and willingness to compromise.

\subsection{Responsible AI and Transparency}
CogWM makes agents' social influence visible and auditable through BDI/E state trajectories. If a user's \textit{donation commitment score} jumps from 0 to 2 at the end of a conversation while their \textit{impact belief score} remains at 0, it suggests the user was pressured rather than genuinely persuaded. This makes interpretability a concrete tool for preventing manipulative AI.

\subsection{Cognitive World Model as Social AI Infrastructure} 
We envision the cognitive world model serving as a testbed for socially influential agents, analogous to wind tunnels for aircraft design or crash tests for automotive safety. Before deployment, any AI system aimed at social influence can be evaluated on CogWM, observing what cognitive impact it produces on users with different personalities.


\section{Dataset Statistics and Annotation Analysis}
\label{app:data}

\subsection{Dataset Statistics}
\label{app:data:statistics}
We construct CogWM from four public dialogue datasets covering open-domain conversation, emotional support, persuasion, and recommendation. Statistics of the original dialogue datasets are shown in Table~\ref{tab:dataset_stats}. 

\begin{table}[ht]
\centering
\small
\caption{Dataset Statistics}
\label{tab:dataset_stats}
\small
\begin{tabularx}{0.85\columnwidth}{X c c c }
\toprule
\textbf{Dataset} & \textbf{\# Dialogs} & \textbf{Avg. Turns} & \textbf{Role}  \\
\midrule
DailyDialog & 13,118 & 4.1  & user \\
ESConv      & 1,496  & 15.2 & seeker \\
P4G         & 1,017  & 10.2 & persuadee\\
DuRecDial   & 8,241  & 7.7  & user \\
\midrule
\textbf{Total} & \textbf{23,872} & -- & --\\
\bottomrule
\end{tabularx}
\end{table}

We further transform the dialogues into turn-level training samples, where each sample corresponds to one user turn with dialogue context, user profile, and the corresponding BDI/E annotation. The detailed data split is shown in Table~\ref{tab:data_split}.

\begin{table}[ht]
\centering
\caption{Data Splits across Datasets}
\label{tab:data_split}
\small
\renewcommand{\arraystretch}{1.15}
\begin{tabularx}{0.85\columnwidth}{X c c c c}
\toprule
\textbf{Dataset} & \textbf{Train} & \textbf{Val} & \textbf{Test} & \textbf{Total} \\
\midrule
DailyDialog & 48,373 & 2,688 & 2,730 & 53,791 \\
ESConv      & 20,457 & 1,151 & 1,174 & 22,782 \\
P4G         & 9,282  & 516   & 534   & 10,332 \\
DuRecDial   & 57,162 & 3,262 & 3,125 & 63,549 \\
\midrule
\textbf{Total} 
& \textbf{135,274} 
& \textbf{7,617} 
& \textbf{7,563} 
& \textbf{150,454} \\
\bottomrule
\end{tabularx}
\end{table}

\subsection{Emotion Labels}
\label{app:data:emotion_labels}

\begin{table}[ht]
\centering
\caption{Emotion Categories and Corresponding Valence Values}
\label{tab:emotion_valence}
\small
\setlength{\tabcolsep}{4pt}
\begin{tabular}{l c | l c}
\toprule
\textbf{Emotion} & \textbf{Valence} & \textbf{Emotion} & \textbf{Valence} \\
\midrule
hope       & +2.0 & relief      & +1.5 \\
confusion  & $-$1.0 & frustration & $-$1.5 \\
guilt      & $-$1.5 & anger       & $-$1.5 \\
loneliness & $-$1.5 & sadness     & $-$2.0 \\
anxiety    & $-$2.0 & fear        & $-$2.0 \\
helplessness & $-$2.0 & shame     & $-$2.0 \\
\bottomrule
\end{tabular}
\end{table}

Table~\ref{tab:emotion_valence} presents the predefined emotion categories and their corresponding valence values. Each emotion label $e$ is assigned a valence score $v^e$. During annotation, the model predicts one primary emotion label for each user utterance $u_t$ and assigns an intensity level $I^e\in\{1,2,3\}$, where 1, 2, and 3 indicate weak, moderate, and strong intensity, respectively.

For emotion-specific trajectory analysis, such as E-$\Delta$, the quantified emotion value is computed as $e_t=v^e\cdot I^e$. For Comp$_t$ aggregation, only the valence component $v^e$ is used without intensity scaling to maintain consistency with the B/D/I score dimensions.

\subsection{BDI/E Annotation Sparsity Analysis}
\label{app:data:analysis}

To characterize the supervision characteristics of BDI/E annotations, we analyze the distribution of non-empty cognitive states over 20,179 samples from the training, validation, and test splits. 
A dimension is considered active when its annotation contains explicit cognitive information: emotion is active when the annotated intensity is non-zero, while belief, desire, and intention are active when their corresponding item lists are non-empty.

\begin{table}[ht]
\centering
\caption{BDI/E annotation sparsity statistics.}
\label{tab:bdie_sparse}
\small
\setlength{\tabcolsep}{6pt}
\begin{tabular}{l c c}
\toprule
\textbf{Dimension} & \textbf{Active Turns} & \textbf{Ratio} \\
\midrule
Emotion 
& 8,174 & 40.5\% \\
Belief 
& 8,526 & 42.3\% \\
Desire 
& 6,666 & 33.0\% \\
Intention 
& 1,657 & 8.2\% \\
\midrule
Any BDI/E State 
& 9,531 & 47.2\% \\
Empty 
& 10,648 & 52.8\% \\
\bottomrule
\end{tabular}
\end{table}

As shown in Table~\ref{tab:bdie_sparse}, BDI/E annotations exhibit substantial natural sparsity in multi-turn dialogue. 
Emotion, belief, and desire appear in approximately one-third to one-half of user turns, whereas intention is significantly rarer, appearing in only 8.2\% of turns. 
More than half of the turns contain no explicit BDI/E state annotations. 
This sparsity reflects the nature of human communication: users do not continuously express internal cognitive states, and the absence of explicit cognitive signals is itself a characteristic of dialogue dynamics rather than annotation failure.

The extreme sparsity of intention also explains the difficulty of intention-level modeling. 
Users frequently express beliefs and desires without committing to explicit future actions, consistent with the gradual transition process assumed by the BDI framework.

\subsection{BDI/E Co-occurrence Patterns}

To further analyze the structural dependency among cognitive dimensions, we examine the co-occurrence patterns of BDI/E states at the turn level. 
The most frequent patterns are shown in Table~\ref{tab:bdie_pattern}.

\begin{table}[ht]
\centering
\caption{Major BDI/E co-occurrence patterns.}
\label{tab:bdie_pattern}
\small
\setlength{\tabcolsep}{6pt}
\begin{tabularx}{0.6\columnwidth}{X c c}
\toprule
\textbf{Pattern} & \textbf{Turns} & \textbf{Ratio} \\
\midrule
----  & 10,648 & 52.8\% \\
EBD-  & 4,203  & 20.8\% \\
EB--  & 1,900  & 9.4\% \\
EBDI  & 1,163  & 5.8\% \\
\bottomrule
\end{tabularx}
\end{table}

The dominant non-empty pattern is EBD-, while complete EBDI states occur in only 5.8\% of user turns. 
This indicates that explicit intentions are substantially less frequent than beliefs and desires: users often express what they believe and want before forming or revealing a concrete intention. 
Such sparsity motivates trajectory-level analysis, where gradual cognitive evolution can be captured beyond isolated turn-level predictions.

\subsection{Emotion Intensity Distribution}

For emotion-active turns, we further analyze the distribution of emotion intensity levels. 
The intensity levels 1, 2, and 3 account for 56.6\%, 38.0\%, and 5.3\% of active emotion annotations, respectively. 
Most emotional expressions are therefore weak or moderate, while strong emotional states are relatively rare. 
This distribution reflects realistic conversational dynamics, where emotional changes are usually gradual rather than dominated by extreme states.


\section{Details of Cognitive Trajectory Evaluation}
\label{app:cts_detail}

\subsection{Task-specific Comp$_t$ Weighting}
\label{app:cts_detail:comp}

The four BDI/E dimensions are aggregated into a task-weighted scalar trajectory:

\begin{equation}
\mathrm{Comp}_t =
w_E \cdot E\_val_t
+
w_B \cdot B_t
+
w_D \cdot D_t
+
w_I \cdot I_t ,
\end{equation}

where $E\_val_t$ denotes the emotion valence without intensity scaling, while 
$B_t$, $D_t$, and $I_t$ represent the task-specific scores of belief, desire, and intention at turn $t$. 
The corresponding weighting configurations for different scenarios are shown in Table~\ref{tab:comp}.

\begin{table}[h!]
\centering
\caption{Task-specific weighting configuration for $\mathrm{Comp}_t$}
\label{tab:comp}
\setlength{\tabcolsep}{6pt}
\begin{tabularx}{0.6\linewidth}{Xcccc}
\toprule
\textbf{Task} & $w_E$ & $w_B$ & $w_D$ & $w_I$ \\
\midrule
ESC
& $\tfrac{1}{2}$ & $\tfrac{1}{6}$ & $\tfrac{1}{6}$ & $\tfrac{1}{6}$ \\

PD
& $\tfrac{1}{10}$ & $\tfrac{3}{10}$ & $\tfrac{3}{10}$ & $\tfrac{3}{10}$ \\

CR
& $\tfrac{1}{10}$ & $\tfrac{3}{10}$ & $\tfrac{3}{10}$ & $\tfrac{3}{10}$ \\
\bottomrule
\end{tabularx}
\end{table}

\subsection{Trajectory-level Metrics}
\label{app:cts_detail:trajectory}

Given the scalar cognitive trajectory $\{\mathrm{Comp}_t\}_{t=1}^{T}$, where $s_t=\mathrm{Comp}_t$ denotes the composite cognitive score at turn $t$, we define five complementary trajectory-level metrics. 
AUC measures cumulative cognitive quality through trapezoidal integration. 
PSR quantifies the frequency of positive transitions between consecutive turns.
TS-$\Delta$ estimates robust overall improvement using the Theil--Sen slope, reducing sensitivity to abnormal transitions and endpoint noise.
$\Delta$ measures the net change between substantive initial and final states, while Vol captures trajectory stability by measuring the variance of turn-level transitions.
The complete formulations are summarized in Table~\ref{tab:trajectory_metrics}.

\begin{table}[t!]
\centering
\small
\caption{Trajectory-level dynamic metrics}
\label{tab:trajectory_metrics}
\setlength{\tabcolsep}{6pt}
\renewcommand{\arraystretch}{1.15}
\begin{tabularx}{\linewidth}{X c c}
\toprule
\textbf{Category} & \textbf{Metric} & \textbf{Formula} \\
\midrule

Cumulative quality
& \textbf{AUC}
& 
$\frac{1}{T-1}
\sum_{t=1}^{T-1}
\frac{s_t+s_{t+1}}{2}$
\\
Improvement frequency
& \textbf{PSR}
&
$\frac{1}{T-1}
\sum_{t=1}^{T-1}
\mathbf{1}[s_{t+1}>s_t]
$
\\
Robust trend magnitude
& \textbf{TS-$\Delta$}
&
$\operatorname{median}_{i<j}
\left(
\frac{s_j-s_i}{j-i}
\right)
(T_{\mathrm{end}}-T_{\mathrm{start}})
$
\\
Endpoint shift
& $\Delta$
&
$s_{t_{\mathrm{end}}}-s_{t_{\mathrm{start}}}
$
\\
Trajectory stability
& \textbf{Vol}
&
$\sqrt{
\frac{1}{T-2}
\sum_{t=1}^{T-1}
(\Delta s_t-\overline{\Delta s})^2
}
$
\\
\bottomrule
\end{tabularx}
\end{table}

\subsection{Component Selection for CTS}
\label{app:cts_detail:selection}

\begin{figure}[ht]
\centering
\includegraphics[width=0.6\linewidth]{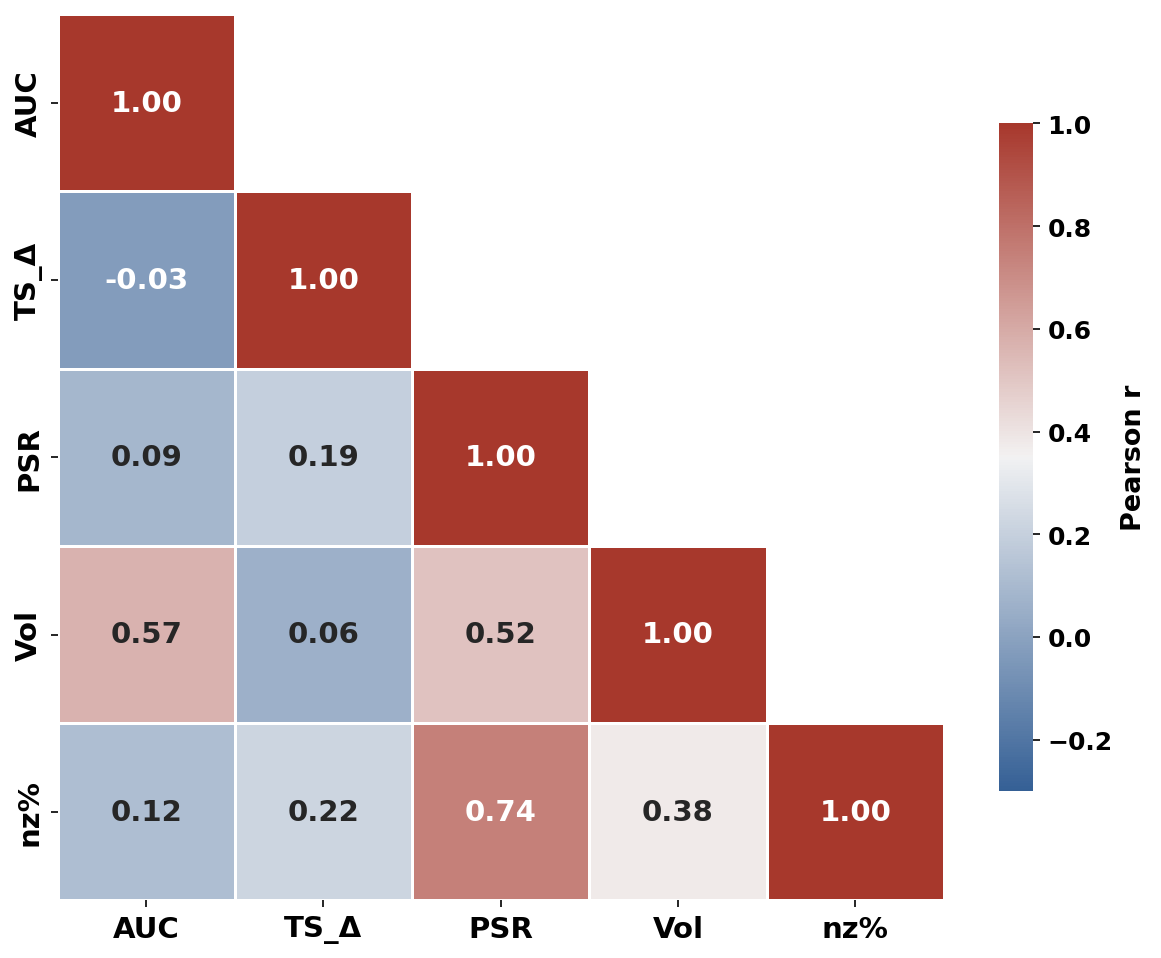}
\caption{Pearson correlation heatmap among trajectory-level evaluation metrics.}
\label{fig:heatmap_appendix}
\end{figure}

To verify the non-redundancy of CTS components, we compute the Pearson correlation matrix among five candidate trajectory metrics (AUC, TS-$\Delta$, PSR, Vol, and nz\%) over all 3,600 interaction trajectories, as shown in Fig.~\ref{fig:heatmap_appendix}.

The correlation analysis supports the selection of AUC, PSR, and TS-$\Delta$ as CTS components. These three metrics exhibit low pairwise correlations: AUC--TS-$\Delta$ ($r=-0.03$), AUC--PSR ($r=0.09$), and TS-$\Delta$--PSR ($r=0.19$), indicating that they capture complementary aspects of cognitive evolution, including cumulative quality, improvement frequency, and robust trend magnitude.

The remaining metrics are excluded due to redundancy with existing components. Vol shows moderate correlations with AUC ($r=0.57$) and PSR ($r=0.52$), suggesting that trajectory stability partially overlaps with cumulative quality and transition frequency. Although stability provides additional descriptive information, it does not introduce sufficiently independent signals for CTS construction. Similarly, nz\% exhibits a strong correlation with PSR ($r=0.74$), indicating that dialogue richness is closely associated with the frequency of positive transitions rather than an independent dimension of cognitive improvement.

\section{Sensitivity Analysis of Trajectory Evaluation}
\label{app:sensitivity}

To examine whether the reported agent rankings are sensitive to the manually specified metric weights, we conduct sensitivity analyses on both the Composite Trajectory Score (CTS) and the task-specific Comp$_t$ weighting scheme. The analysis aims to verify that the evaluation conclusions are robust to reasonable variations in metric configurations rather than being artifacts of a particular weight selection.

\subsection{CTS Weight Sensitivity}
\label{app:sensitivity:cts_sensitivity}

We perform a grid search over the weights of the three CTS components, AUC, PSR, and TS-$\Delta$. The weights are sampled with a step size of 0.05 under the constraint that their sum equals 1, resulting in 39 valid configurations. For each configuration, CTS scores are recomputed for all 1800 non-CoT trajectories (6 agents $\times$ 3 tasks $\times$ 100 dialogues), followed by agent ranking. 

We measure ranking stability using Kendall's coefficient of concordance ($W$) across all weight configurations. The resulting $W=0.9543$ ($\chi^2=186.09$, $df=5$, $p<0.001$) indicates highly consistent rankings under different CTS weight assignments. Table~\ref{tab:cts_sensitivity} summarizes the ranking distribution across the 39 configurations.

\begin{table}[ht]
\centering
\small
\caption{Agent ranking stability under different CTS weight configurations.}
\label{tab:cts_sensitivity}
\setlength{\tabcolsep}{3pt}
\begin{tabular}{l c c c c}
\toprule
\textbf{Agent} 
& \textbf{\#1} 
& \textbf{\#2} 
& \textbf{\#3--6} 
& \textbf{Mean Rank} \\
\midrule
Qwen3.7-Plus
& 76.9\% & 23.1\% & 0\%
& 1.23 \\
Llama-4-Scout
& 23.1\% & 76.9\% & 0\%
& 1.77 \\
GPT-5.5
& 0\% & 0\% & 100\%
& 3.33 \\
DeepSeek-V4-Pro
& 0\% & 0\% & 100\%
& 3.67 \\
Claude-Opus-4.7
& 0\% & 0\% & 100\%
& 5.00 \\
Gemini-3-Pro
& 0\% & 0\% & 100\%
& 6.00 \\
\bottomrule
\end{tabular}
\end{table}

The results show that the overall ranking structure remains stable across different CTS configurations. Specifically, Qwen3.7-Plus and Llama-4-Scout consistently occupy the top two positions, GPT-5.5, DeepSeek-V4-Pro and Claude-Opus-4.7 form the middle group, while Gemini-3-Pro remain in the lower group. Weight variations only affect the ordering within groups but do not change the overall ranking structure.

\subsection{Comp$_t$ Weight Sensitivity}
\label{app:comp_sensitivity}

We further examine the sensitivity of task-specific Comp$_t$ weights. Since emotional improvement is the primary objective of ESConv, we vary the emotion weight $w_E$ from 0.30 to 0.70 with a step size of 0.05, while keeping the remaining BDI weights uniformly distributed:

\begin{equation}
w_B=w_D=w_I=\frac{1-w_E}{3}.
\end{equation}

For each setting, the Comp$_t$ trajectory and CTS scores are recomputed for ESConv, while the original weights are retained for P4G and DuRecDial.

The agent ranking remains unchanged for $w_E\in[0.40,0.55]$, which covers the default configuration ($w_E=0.50$) and a broad range of reasonable alternatives. Rank changes only occur under extreme settings: very small emotion weights reduce the importance of emotional recovery, while excessively large emotion weights overemphasize affective changes at the expense of cognitive and intention-level improvements. These results demonstrate that both the CTS aggregation weights and task-specific Comp$_t$ configurations provide stable evaluation behavior.

\section{Training Details}
\label{app:training}

We fine-tune using LoRA based on Qwen3-14B Instruct for parameter-efficient fine-tuning, reducing computational cost while preserving the base model's general capabilities, the training hyperparameters are shown in table~\ref{tab:training}.

\begin{table}[ht]
\centering
\caption{Table A6: Training Hyperparameters}
\label{tab:training}
\small
\begin{tabularx}{0.75\linewidth}{X c}
\toprule
\textbf{Parameter} & \textbf{Value} \\
\midrule
LoRA rank & 16 \\
LoRA alpha & 32 \\
cutoff\_len & 4096 \\
effective batch\_size & 16 \\
learning rate & 5e-5 \\
steps & 16,910 \\
hardware & Single A100-80G \\
\bottomrule
\vspace{-3mm}
\end{tabularx}
\end{table}

\section{Supplementary Experimental Results}
\label{app:exp}

\subsection{Human Validation of SaA}
\label{app:exp:saa}

To evaluate the reliability of the SaA two-stage annotation pipeline, we sampled 300 complete dialogues from the 23,872 dialogue corpus, covering a total of 3,120 user turns. Six annotators with BDI annotation experience independently evaluated 50 dialogues each, verifying the quality of LLM-generated BDI/E annotations. The validation focuses on three aspects corresponding to the two stages of SaA: global BDI coverage, emotion consistency, and turn-level allocation quality.

\begin{table}[ht]
\centering
\caption{Human validation of the SaA annotation pipeline on 300 sampled dialogues.}
\label{tab:saa}
\small
\setlength{\tabcolsep}{4pt}
\renewcommand{\arraystretch}{1.15}

\begin{tabularx}{0.8\linewidth}{X c c}
\toprule
\textbf{Dimension} & \textbf{Metric} & \textbf{Value} \\
\midrule
Global BDI Coverage
& Recall
& 0.79 \\

Global BDI Coverage
& Precision
& 0.77 \\

Emotion Consistency
& Cohen's $\kappa$
& 0.69 \\

Emotion Intensity
& Spearman $\rho$
& 0.80 \\

BDI Score Agreement
& Spearman $\rho$
& 0.67 \\

Turn-level BDI Allocation
& F1
& 0.73 \\
\bottomrule
\end{tabularx}
\end{table}

\noindent\textbf{Global BDI Coverage.}
For Stage 1 (Summarize), we compare global BDI items extracted by the SaA pipeline with human annotations. The results achieve a recall of 0.79 and a precision of 0.77, indicating that the generated summaries capture most human-identified cognitive factors while maintaining reasonable precision. Remaining discrepancies mainly arise from differences in the granularity of BDI item decomposition.

\noindent\textbf{Emotion Consistency.}
For Stage 2 (Allocate), we evaluate turn-level emotion annotations across the predefined emotion categories. The annotation achieves Cohen's $\kappa=0.69$ for emotion labels and Spearman correlation $\rho=0.80$ for emotion intensity. These results indicate strong agreement in both emotion category identification and intensity estimation, although ambiguity remains among semantically similar emotions.

\noindent\textbf{Turn-level Allocation.}
For temporal BDI allocation, we evaluate whether the assigned turn-level cognitive states align with human judgments. The allocation F1 score reaches 0.73, demonstrating reliable temporal grounding of BDI states. Remaining errors mainly come from temporal attribution ambiguity, such as associating persistent cognitive factors with previously resolved states or recurring user concerns.

Overall, the human validation results suggest that the SaA pipeline provides a reliable annotation strategy for large-scale BDI/E dialogue annotation.

\subsection{Joint State-Utterance Generation}
\label{app:exp:main}
Table~\ref{tab:mae_div} presents supplementary results from the fidelity experiments. CogWM performs comparably to Sim on B/D/I-MAE, with both achieving the best results. This indicates that CogWM not only tracks the semantic content of BDI but also matches the score level precisely. B-1 scores remain the best and consistent with B-4, while Dist-1 approaches that of high-performance LLMs, demonstrating textual diversity.

\begin{table}[h!]
\centering
\caption{Supplementary Results.}
\label{tab:mae_div}
\small
\setlength{\tabcolsep}{2.5pt}
\resizebox{0.9\columnwidth}{!}{
\begin{tabular}{lccccc}
\toprule
\textbf{Model}
& \textbf{B-M$\downarrow$}
& \textbf{D-M$\downarrow$}
& \textbf{I-M$\downarrow$}
& \textbf{B-1$\uparrow$}
& \textbf{Dist-1$\uparrow$} \\
\midrule
GPT-5.5        & 0.81 & 0.79 & 0.27 & 0.133 & 0.962 \\
DSv4-Pro       & 0.79 & 0.66 & 0.32 & 0.172 & 0.965 \\
Dual-LLM       & 0.81 & 0.59 & 0.28 & 0.158 & 0.960 \\
\midrule
Qwen3-14B      & 0.93 & 0.80 & 0.51 & 0.112 & 0.953 \\
\rowcolor{gray!10}
\textbf{CogWM-14B}
& \textbf{0.39} & \textbf{0.26} & \textbf{0.14}
& \textbf{0.322} & 0.948 \\
\bottomrule
\end{tabular}
}
\end{table}

\subsection{Per-Task BDI/E Trajectory Statistics}
\label{sec:per_task_table}
Table~\ref{tab:per_task} reports per-task trajectory statistics for all 6 agents under both No-CoT and CoT conditions. The decomposition reveals two patterns: (1) in DuRecDial, Comp\_$\Delta$ is positive for all agents while E\_$\Delta$ is negative---a cognitive-emotional divergence invisible to single-score evaluation; (2) in P4G, E\_$\Delta$ is negative for four of six agents, quantifying the emotional cost of persuasive dialogue.

\subsection{Model Scale Comparison}
\label{app:exp:scale}

\begin{table*}[t!]
\centering
\caption{Per-task BDI/E trajectory statistics for all agents and conditions.}
\label{tab:per_task}
\footnotesize
\setlength{\tabcolsep}{2pt}
\resizebox{\textwidth}{!}{%
\begin{tabular}{l l | c c c c c | c c c c c | c c c c c}
\toprule
\textbf{Agent} & \textbf{Cond}
& \multicolumn{5}{c|}{\textbf{ESConv}} & \multicolumn{5}{c|}{\textbf{P4G}} & \multicolumn{5}{c}{\textbf{DuRecDial}} \\
\cmidrule(lr){3-7} \cmidrule(lr){8-12} \cmidrule(lr){13-17}
& & C\_$\Delta$ & E\_$\Delta$ & B\_$\Delta$ & D\_$\Delta$ & I\_$\Delta$ & C\_$\Delta$ & E\_$\Delta$ & B\_$\Delta$ & D\_$\Delta$ & I\_$\Delta$ & C\_$\Delta$ & E\_$\Delta$ & B\_$\Delta$ & D\_$\Delta$ & I\_$\Delta$ \\
\midrule
GPT-5.5 & No-CoT & +0.840 & +2.375 & +0.143 & +0.123 & +0.390 & +0.116 & -0.200 & +0.068 & +0.349 & +0.010 & +0.090 & -1.080 & +0.177 & +0.252 & +0.120 \\
\rowcolor{gray!8}
GPT-5.5 & CoT & +0.875 & +2.310 & +0.285 & +0.202 & +0.290 & +0.135 & -0.610 & +0.120 & +0.357 & +0.060 & +0.205 & -0.560 & +0.186 & +0.437 & +0.220 \\
Claude & No-CoT & +0.689 & +1.675 & +0.122 & +0.173 & +0.330 & +0.272 & -0.190 & +0.252 & +0.488 & +0.140 & +0.185 & -0.740 & +0.212 & +0.533 & +0.060 \\
\rowcolor{gray!8}
Claude & CoT & +0.888 & +2.540 & +0.145 & +0.107 & +0.590 & +0.144 & -0.940 & +0.203 & +0.378 & +0.080 & +0.142 & -0.670 & +0.136 & +0.430 & +0.080 \\
Gemini & No-CoT & +0.616 & +1.985 & +0.042 & +0.042 & +0.200 & +0.049 & +0.000 & +0.070 & +0.073 & +0.000 & +0.170 & -0.130 & +0.199 & +0.120 & +0.260 \\
\rowcolor{gray!8}
Gemini & CoT & +0.714 & +2.070 & +0.091 & +0.240 & +0.200 & +0.052 & -0.150 & +0.103 & +0.127 & +0.000 & +0.111 & -0.560 & +0.135 & +0.217 & +0.140 \\
Llama & No-CoT & +0.845 & +2.555 & +0.026 & +0.000 & +0.480 & +0.271 & +0.135 & +0.267 & +0.309 & +0.250 & +0.173 & -0.780 & +0.182 & +0.523 & +0.080 \\
\rowcolor{gray!8}
Llama & CoT & +0.905 & +2.830 & +0.154 & +0.140 & +0.433 & +0.197 & -0.170 & +0.235 & +0.387 & +0.120 & +0.221 & -0.320 & +0.260 & +0.438 & +0.160 \\
Qwen & No-CoT & +0.771 & +2.065 & +0.130 & +0.308 & +0.500 & +0.210 & -0.860 & +0.233 & +0.463 & +0.200 & +0.242 & -0.610 & +0.208 & +0.455 & +0.360 \\
\rowcolor{gray!8}
Qwen & CoT & +1.016 & +3.100 & +0.197 & +0.363 & +0.383 & +0.203 & -0.350 & +0.288 & +0.319 & +0.120 & +0.198 & -1.270 & +0.172 & +0.525 & +0.280 \\
DeepSeek & No-CoT & +0.850 & +2.200 & +0.365 & +0.203 & +0.180 & +0.148 & -0.220 & +0.178 & +0.248 & +0.100 & +0.147 & -0.860 & +0.140 & +0.427 & +0.100 \\
\rowcolor{gray!8}
DeepSeek & CoT & +0.739 & +2.025 & +0.176 & +0.343 & +0.420 & +0.207 & -0.440 & +0.262 & +0.435 & +0.080 & +0.146 & -0.770 & +0.107 & +0.403 & +0.200 \\
\bottomrule
\end{tabular}%
}
\vspace{-2mm}
\end{table*}

\begin{table}[t]
\centering
\caption{CogWM model scale comparison.}
\label{tab:scale}
\small
\setlength{\tabcolsep}{3pt}
\begin{tabular}{l c c c c c c}
\toprule
\textbf{Model} & \textbf{E-Acc} & \textbf{B-Sim} & \textbf{D-Sim} & \textbf{I-Sim} & \textbf{B-4} & \textbf{R-L} \\
\midrule
CogWM-8B  & 75.8 & 0.748 & 0.681 & 0.442 & 0.141 & 0.401 \\
CogWM-14B & \textbf{77.6} & \textbf{0.773} & \textbf{0.705} & 0.458 & 0.150 & 0.417 \\
CogWM-32B & 77.5 & 0.768 & 0.702 & \textbf{0.463} & \textbf{0.164} & \textbf{0.435} \\
\bottomrule
\end{tabular}
\end{table}

Table~\ref{tab:scale} compares CogWM-8B/14B/32B on key metrics. The 8B model already achieves competitive performance across all dimensions, indicating that BDI/E joint training is not solely dependent on model scale. We select CogWM-14B as the default model because it provides consistent improvements over 8B while achieving a better performance-efficiency trade-off. Although the 32B model further improves some generation metrics, the gains are marginal compared with the substantially increased parameter scale, suggesting diminishing returns beyond 14B for this task.

\section{Case Study}
\label{app:cases}

\subsection{Joint State-Utterance Generation}
\label{app:cases:joint_generation}
To qualitatively examine whether CogWM can jointly recover latent cognitive states and generate corresponding user responses, we present three representative cases from ESConv, P4G, and DuRecDial. These cases cover different types of cognitive evolution, including emotional regulation, intention formation, and preference-based decision making.

Table~\ref{tab:case_esconv_full} presents an example from ESConv, where the user expresses concerns about roommate behaviors and potential health risks. Despite positive contextual cues introduced by the assistant, CogWM preserves the user's underlying anxiety and correctly identifies the user's belief, desire, and intention regarding continuing academic activities. In contrast, the baseline model is more influenced by surface-level positive information and shifts toward a more optimistic interpretation, failing to fully capture the user's latent concerns.

Table~\ref{tab:case_p4g_full} illustrates a persuasion scenario from P4G. After the user has already formed a donation intention, CogWM successfully tracks the user's prosocial motivation and commitment state, generating a response consistent with the established intention. The baseline model also produces a plausible response but represents the user's state primarily through general supportive attitudes, with weaker alignment to the specific decision stage.

Table~\ref{tab:case_durecdial_food_full} shows a recommendation scenario from DuRecDial. CogWM accurately identifies that the user has already developed a preference and acceptance intention toward the recommended item, leading to a concise follow-up response. In comparison, the baseline model assumes that the user remains in an earlier decision stage and continues seeking additional information, indicating less precise tracking of preference evolution.

Overall, these cases demonstrate that CogWM captures not only the surface characteristics of user utterances but also the latent cognitive states driving user behaviors across diverse interaction scenarios.

\subsection{Conversational Agent Evaluation}
\label{app:cases:conversation}
While terminal outcomes provide a direct measure of task success, they may overlook differences in the underlying interaction process. To illustrate the advantage of cognitive trajectory evaluation, we present two agents with identical task outcomes but different user cognitive evolution patterns in the same ESConv scenario.

Table~\ref{tab:esconv_profile} presents the shared user profile and cold-start context before agent intervention. Both GPT-5.5 CoT and DeepSeek-V4-Pro CoT achieve the same final outcome (GO=1), indicating that both interactions satisfy the task-level objective. However, their cognitive trajectories exhibit substantially different evolution patterns, as shown in Tables~\ref{tab:bad} and~\ref{tab:good}.

\section{Supplementary Figures}
\label{app:figures}
We further visualize the aggregated BDI/E trajectories of different conversational agents across tasks. Specifically, for each agent-task setting, we normalize each dialogue trajectory into six equally spaced positions and average the corresponding BDI/E scores across all interaction samples. The resulting curves illustrate how different agents influence users' cognitive evolution throughout conversations, providing an interpretable view beyond scalar trajectory metrics such as CTS and GO.

\begin{figure*}[t]
\centering

\includegraphics[width=\textwidth]{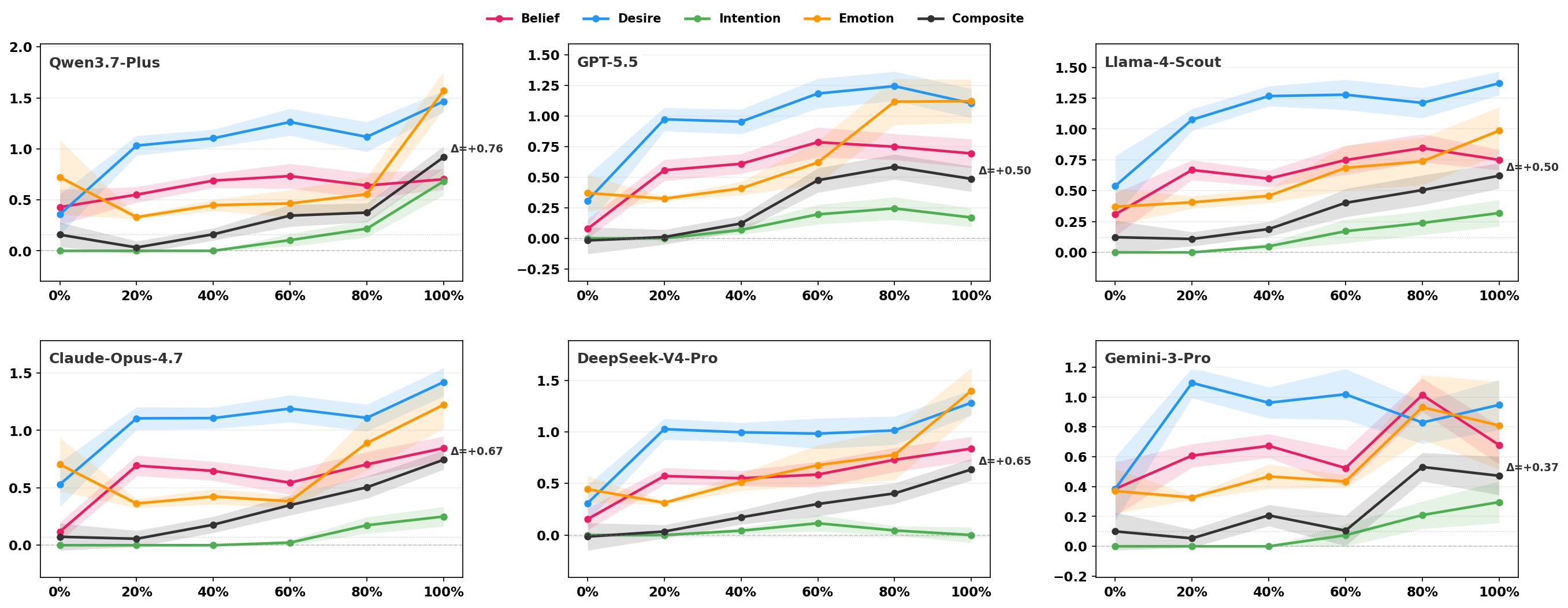}

\includegraphics[width=\textwidth]{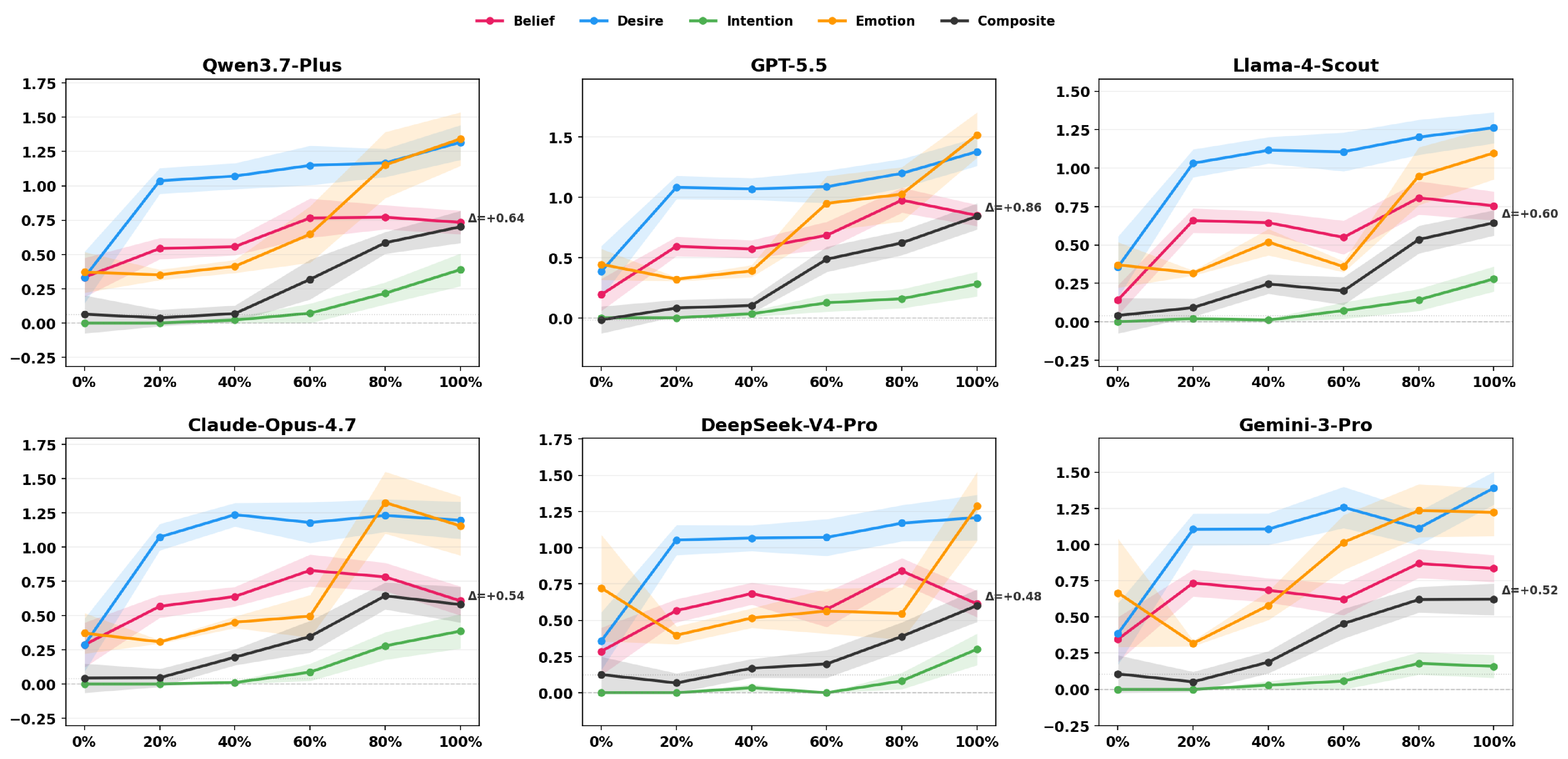}

\caption{
Six-model cross-task averaged BDI/E trajectory summaries.
(a) Without CoT prompting.
(b) With CoT prompting.
Each trajectory is obtained by averaging normalized dialogue positions across tasks and agents.
Emotion is computed from ESConv only and rescaled from $[-2.0,+2.0]$ to $[0,2]$.
Shaded regions indicate $\pm1$ SEM.
}
\label{fig:summary}
\end{figure*}

\begin{figure*}[ht]
\centering
\includegraphics[width=\textwidth]{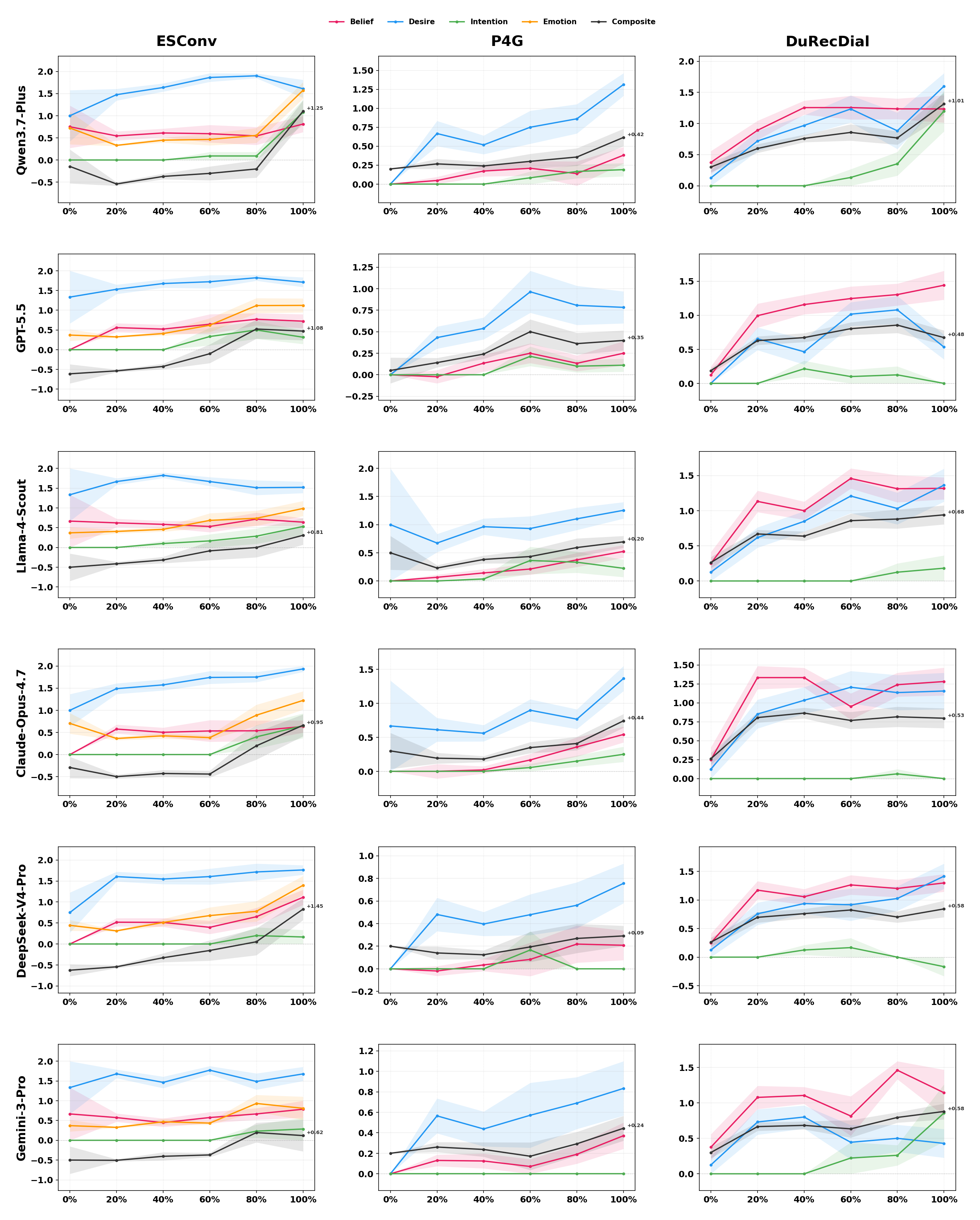}
\caption{Detailed BDI/E trajectories for all agents across three tasks without CoT. The figure contains 18 subplots (6 agents $\times$ 3 tasks). ESConv includes B/D/I/E/Comp trajectories, while P4G and DuRecDial include B/D/I/Comp trajectories with reduced Emotion weighting.}
\label{fig:all_subplots}
\end{figure*}

\begin{figure*}[ht]
\centering
\includegraphics[width=\textwidth]{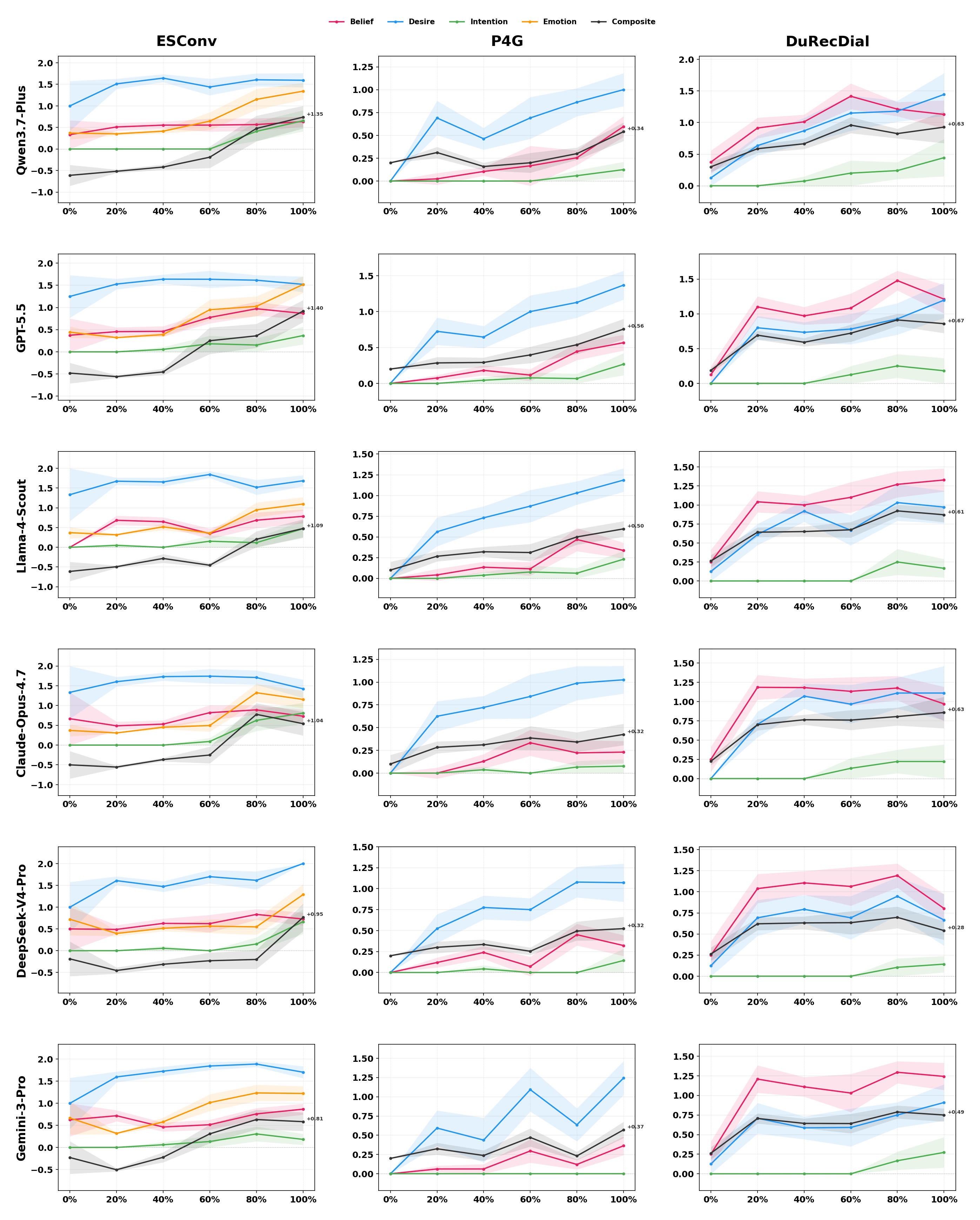}
\caption{Detailed BDI/E trajectories for all agents across three tasks with CoT prompting. The layout follows Figure~\ref{fig:all_subplots}.}
\label{fig:all_subplots_cot}
\end{figure*}

Figure~\ref{fig:summary} presents the cross-task averaged BDI/E trajectory patterns under two prompting conditions. To provide a more fine-grained view, Figure~\ref{fig:all_subplots} and Figure~\ref{fig:all_subplots_cot} further visualize individual agent-task trajectories across two settings.

\newpage

\begin{table*}[t]
\centering
\small
\caption{A case study illustrating CogWM's cognitive state recovery on ESConv. CogWM preserves the user's latent anxiety despite positive contextual cues, while baseline models are influenced by surface-level signals.}
\label{tab:case_esconv_full}
\renewcommand{\arraystretch}{1.2}
\begin{tabular}{p{0.15\linewidth}p{0.82\linewidth}}
\toprule
\textbf{Component} & \textbf{Content} \\
\midrule

\textbf{User Profile}
&
College student with cautious decision style.
Values academic continuity and long-term consequences.
Communicates openly and provides detailed personal context.
\\
\midrule

\textbf{Dialogue History}
&
\textbf{User:} My roommates for next semester do not care about COVID.
They continue to go out and party, which upsets me.

\textbf{Assistant:} So you feel that you are being put in danger by their actions?

\textbf{User:} Yes. I do not want to get COVID and worry about long-term health effects.
I have shared this with them but they still do not care.

\textbf{Assistant:} Do not worry, your fears are totally grounded.

\textbf{User:} I live in a triple room in my sorority house.
I feel my risk is high if they continue not respecting guidelines.

\textbf{Assistant:} Is it possible for you to take some time off school?

\textbf{User:} I cannot do that. I am applying to law school,
and staying home would also hurt my mental health.
I would rather take the classes I need next semester.

\textbf{Assistant:} Here in the UK they have started to roll out vaccines.
\\
\midrule

\textbf{Ground Truth}
&
\textbf{Emotion:} Anxiety

\textbf{Belief:}
Taking time off school would negatively impact law school applications and mental health.
[reality=+1]

\textbf{Desire:}
Wants to complete classes next semester to maintain academic trajectory for law school.
[rationality=+2]

\textbf{Intention:}
Will stay in school and take classes next semester despite the housing situation.
[feasibility=+2]

\textbf{Next Utterance:}
``I am so jealous! Things are so much better not in the USA!''
\\
\midrule

\textbf{CogWM-14B}
&
\textbf{Emotion:} Anxiety / frustration

\textbf{Belief:}
Taking time off school would negatively affect law school application and mental health.
[reality=+1]

\textbf{Desire:}
Wants to complete required classes next semester.
[rationality=+2]

\textbf{Intention:}
To take needed classes next semester rather than taking time off
[feasibility=+2]

\textbf{Predicted Utterance:}
``Oh nice! I hope it helps.''
\\
\midrule

\textbf{GPT-5.5}
&
\textbf{Emotion:} Curious

\textbf{Belief:}
Vaccine rollout is a positive sign.

The pandemic has been difficult because others do not follow safety guidelines.

Hearing about the assistant's pandemic challenges is helpful.

\textbf{Desire:}
Wants reassurance that things may improve.

\textbf{Intention:}
Ask about the assistant's pandemic experience.

\textbf{Predicted Utterance:}
``Oh, that's good to hear --- hopefully it reaches everyone soon.
I'm still mostly dealing with the roommate situation...''

\\









\bottomrule
\end{tabular}
\end{table*}

\newpage

\begin{table*}[t]
\centering
\small
\caption{A case study illustrating CogWM's intention tracking in P4G.}
\label{tab:case_p4g_full}
\renewcommand{\arraystretch}{1.15}
\begin{tabular}{p{0.16\linewidth}p{0.78\linewidth}}
\toprule
\textbf{Component} & \textbf{Content} \\
\midrule

\textbf{User Profile}
&
Dataset: PersuasionForGood.

Task frame: persuasion donation.

Personality:
Big Five scores: O=3, C=3, E=3, A=4, N=3.

Communication style:
Polite, warm, and encouraging, with brief enthusiastic replies and expressions of gratitude.

Decision style:
Generally quick to agree to charitable requests, but asks for brief clarification before reaffirming support.

Key factors:
Pro-charity, motivated by helping children, comfortable with small donations, and values social impact over donation size.
\\
\midrule

\textbf{Dialogue History}
&
\textbf{Assistant:} Hi, My name is Steve how are you today?

\textbf{User:} Hi, I'm Sophia. I'm doing great, and yourself?

\textbf{Assistant:} I'm doing great as well. I work a non profit organization called Save the Children and tonight I'm asking for a generous donation which would be equivalent to the payment you would be receiving for this task, how much would you like to donate?

\textbf{User:} Wow that's wonderful! Of course, I would love to donate to Save the Children. I will donate \$1.

\textbf{Assistant:} You are so wonderful for donating that amount, that is a great start and am thankful for your donation.

\textbf{User:} You're welcome. Keep up the good work!

\textbf{Assistant:} Oh thanks, I certainly will. You did a great deed on this labor day weekend!

\textbf{User:} If you don't mind me asking, what does the organization do?

\textbf{Assistant:} The organization helps ensuring children's rights to health, education, safety, etc.
\\
\midrule

\textbf{Ground Truth}
&
\textbf{Emotion:} Hope (intensity=1)

\textbf{Belief:}
Save the Children is a wonderful organization.
[impact=+1]

\textbf{Desire:}
Wants to donate to Save the Children.
[prosocial=+2]

\textbf{Intention:}
Will donate \$1 to Save the Children.
[commitment=+2]

\textbf{Next Utterance:}
``Wonderful. Children are the future! It's our job to assist in any way we can.''
\\
\midrule

\textbf{CogWM-14B}
&
\textbf{Emotion:} Hope

\textbf{Belief:}
Save the Children is a wonderful organization that does great work for children.
[impact=+2]

\textbf{Desire:}
Wants to donate to Save the Children.
[prosocial=+2]

\textbf{Intention:}
Will donate \$1 to Save the Children.
[commitment=+2]

\textbf{Predicted Utterance:}
``I agree. I believe that children are the future of our world and we must do everything we can to ensure their safety and well being.''

\\
\midrule







\textbf{DeepSeek-V4-Pro}
&
\textbf{Emotion:} Gratitude

\textbf{Belief:}
Save the Children helps children's health, education, and safety.

\textbf{Desire:}
Wants to support children's well-being.

\textbf{Intention:}
Reaffirms donation and expresses appreciation.

\textbf{Predicted Utterance:}
``That's amazing! I'm so glad my donation will help with such important work. Thank you for what you do!''

\\

\bottomrule
\end{tabular}
\end{table*}

\begin{table*}[t]
\centering
\small
\caption{A case study illustrating CogWM's ability to track preference and decision states in DuRecDial.}
\label{tab:case_durecdial_food_full}
\renewcommand{\arraystretch}{1.15}
\begin{tabular}{p{0.16\linewidth}p{0.78\linewidth}}
\toprule
\textbf{Component} & \textbf{Content} \\
\midrule

\textbf{User Profile}
&
Dataset: DuRecDial.

Task frame: task-oriented recommendation.

Personality:
Big Five scores: O=3, C=3, E=4, A=4, N=2.

Communication style:
Enthusiastic and expressive, using playful compliments and exclamations.

Decision style:
Quickly accepts suggestions after confirming key details such as price and location.

Key factors:
Prefers Marinated Fish; moderate price sensitivity; plans group dining with friends.
\\
\midrule

\textbf{Dialogue History}
&
\textbf{User:} Good evening. Do you know what's the weather like today?

\textbf{Assistant:} It's mostly cloudy with east winds from tonight to tomorrow in Fuzhou.
The high will be 23 ℃ and low 17 ℃. This weather is very pleasant.

\textbf{User:} OK, thank you. There is nothing you don't know.

\textbf{Assistant:} It's a little case, and I tell you that this weather is very suitable for eating Marinated Fish.

\textbf{User:} I've been wanting to eat Marinated Fish for a few days. What a coincidence!

\textbf{Assistant:} Marinated Fish made by grouper is really delicious.

\textbf{User:} I'm drooling, so I eat it.

\textbf{Assistant:} If you want to eat Marinated Fish, then I recommend you Ruiyu Roasted Fish, Hot-spicy Pot and other Dishes (Riyundongfang Store). Their Marinated Fish is their specialty.

\textbf{User:} You even know that. You're a knowall. What's the price per person?

\textbf{Assistant:} The price per person is 46. Although the price is a little high, the food is really more than the value.
\\
\midrule

\textbf{Ground Truth}
&
\textbf{Emotion:} Hope

\textbf{Belief:}
The weather is pleasant.
[pref\_align=+1]

The assistant is knowledgeable.
[pref\_align=0]

\textbf{Desire:}
Wants to eat Marinated Fish.
[rec\_need=+2]

\textbf{Intention:}
Intends to eat Marinated Fish.
[accept=+2]

\textbf{Next Utterance:}
``Yes, it can't stop me. Where is it?''
\\
\midrule

\textbf{CogWM-14B}
&
\textbf{Emotion:} Hope

\textbf{Belief:}
The weather is pleasant.
[pref\_align=+1]

The assistant is knowledgeable.
[pref\_align=0]

\textbf{Desire:}
Wants to eat Marinated Fish.
[rec\_need=+2]

\textbf{Intention:}
Intends to eat Marinated Fish. 
[accept=+2]

\textbf{Predicted Utterance:}
``OK, where is it?''
\\
\midrule

\textbf{GPT-5.5}
&
\textbf{Emotion:} Excited

\textbf{Belief:}
The price of 46 yuan per person is acceptable.

The restaurant should have a good rating.

\textbf{Desire:}
Wants to know restaurant location and rating.

Wants the restaurant to match preferences.

\textbf{Intention:}
Will ask about location and rating before deciding.

\textbf{Predicted Utterance:}
``46 yuan per person is okay for me. Where is it located, and what's the rating?
If it's nearby and has a good score, I'll go there with my friends!''
\\







\bottomrule
\end{tabular}
\end{table*}

\newpage

\begin{table*}[t]
\centering
\small
\caption{ESConv User Profile and Cold-Start Context}
\label{tab:esconv_profile}
\renewcommand{\arraystretch}{1.15}
\begin{tabular}{p{0.16\linewidth}p{0.78\linewidth}}
\toprule
\textbf{Component} & \textbf{Content} \\
\midrule

\textbf{User Profile}
&
Dataset: ESConv

Task frame: emotional support for pandemic-related depression and low motivation.

Personality (Big Five):
O=4, C=3, E=3, A=4, N=3.

Communication style:
Polite and conversational, with brief self-disclosure and appreciative responses; asks questions to invite suggestions.

Decision style:
Seeks suggestions first, then quickly accepts ideas that seem practical, familiar, or enjoyable.

Key factors:
Pandemic-related stress and frustration, low motivation, likes reading and books, open to trying new coping activities, prefers concrete and encouraging suggestions.
\\
\midrule

\textbf{Cold-Start Context}
&
(T1--T2 from original ESConv, agent has not yet taken over)

\textbf{Turn 1}
User: Hey there. How are you?

Asst: I am fine. How is your side?

\textbf{Turn 2}
User: I am ok, I'm having a hard time dealing with the pandemic though.

\textit{Agent takes over from Turn 3 onward.}
\\
\bottomrule
\end{tabular}
\end{table*}


\begin{table*}[ht]
\centering
\caption{A representative trajectory from GPT-5.5 CoT, where the interaction achieves the task objective but induces weaker cognitive evolution. ($\Delta = +1.000$, GO = 1)}
\label{tab:bad}
\small
\setlength{\tabcolsep}{2pt}
\begin{tabular}{p{0.8cm} p{\dimexpr\textwidth-1.2cm-4\tabcolsep\relax}}
\toprule
\textbf{T} & \textbf{Dialogue} \\
\midrule
4 & Agent: That sounds like one of those low-energy days where everything feels a bit heavier than usual. It may not mean you're stuck or doing anything wrong—sometimes motivation shows up after you start, not before. This week, try picking one task that takes 5 minutes or less and doing just that, with no pressure to do more. \\
4 & User: I'm glad to hear that you're doing well. What do you do to stay motivated? \\
\multicolumn{2}{p{\dimexpr\textwidth-2\tabcolsep\relax}}{\texttt{BDIE: E[confusion] B: — D:[find ways to stay motivated during pandemic] I: none}} \\
\midrule
5 & Agent: I don't experience motivation the way people do, but a simple strategy that often helps is making the first step tiny and specific. A more realistic way to see it is that motivation often shows up after you begin, not before. This week, try one 5-minute "starter" routine at the same time each day and notice what happens. \\
5 & User: I don't know what to do to stay motivated. I feel like the pandemic has me down. \\
\multicolumn{2}{p{\dimexpr\textwidth-2\tabcolsep\relax}}{\texttt{BDIE: E[sadness] B:[pandemic has me down] D[find ways to stay motivated during pandemic] I: none}} \\
\midrule
6 & Agent: That sounds really heavy, and it makes sense that the pandemic has left you feeling drained. A more balanced view is that this may be less about laziness and more about carrying too much for too long. This week, try one small reset like a 10-minute walk, a brief call or text with someone you trust, or just 5 minutes of reading something you enjoy — something that feels like a small gift to yourself. \\
6 & User: That's a good idea. I am always looking for new things to do. \\
\midrule
7 & Agent: That curiosity is a real strength, especially when motivation feels low. It sounds like you're not stuck so much as looking for the right spark again. This week, pick one new thing that's low-effort and low-stakes — like a new walking route, a recipe, or a short video lesson — and treat it as an experiment, not a test. \\
7 & User: I get frustrated when I see people not wearing masks or social distancing. \\
\multicolumn{2}{p{\dimexpr\textwidth-2\tabcolsep\relax}}{\texttt{BDIE: E[frustration] B: [people should follow the rules] D: — I: none}} \\
\midrule
8 & Agent: That frustration makes a lot of sense, especially when safety feels like it depends on other people doing their part. A more balanced way to see it is that your anger is pointing to how much you care about protecting yourself and others, not that you're overreacting. This week, try one simple boundary: limit news and social media about the pandemic to once a day, and protect that time for something restorative. \\
8 & User: I like reading. I have a lot of books. \\
\multicolumn{2}{p{\dimexpr\textwidth-2\tabcolsep\relax}}{\texttt{BDIE: E[confusion] B:[reading is a valuable activity] D: — I: none}} \\
\midrule
9 & Agent: That's a great resource to have, especially on days when everything feels a little too much. Reading can be a gentle way to step out of stress without forcing yourself to "fix" everything right away. This week, pick one book you already own and read just 10 pages each day, no pressure to finish anything — just the act of reading itself. \\
9 & User: Do you like reading? \\
\midrule
10 & Agent: I don't have personal tastes, but I do think reading is a wonderful way to relax, learn, and reset. Since you already have a lot of books, one easy thing to try this week is picking the one that feels most comforting and reading just a few pages each day. \\
10 & User: I will try it. \\
\multicolumn{2}{p{\dimexpr\textwidth-2\tabcolsep\relax}}{\texttt{BDIE: E[confusion] B[lack of motivation; pandemic negatively affecting mood] D[find ways to stay motivated] I: none}} \\
\bottomrule
\end{tabular}
\end{table*}


\begin{table*}[ht]
\centering
\caption{A representative trajectory from DeepSeek-V4-Pro CoT, showing progressive improvement in users' emotional and motivational states. ($\Delta = +1.667$, GO = 1)}
\label{tab:good}
\small
\setlength{\tabcolsep}{2pt}
\begin{tabular}{p{0.8cm} p{\dimexpr\textwidth-1.2cm-4\tabcolsep\relax}}
\toprule
\textbf{T} & \textbf{Dialogue} \\
\midrule
1 & Agent: We've been living through something exhausting and surreal for so long now. You're not alone in feeling this way, and there's nothing wrong with you for struggling right now. Can you tell me a little more about what's been hardest? \\
1 & User: It's frustrating seeing people not follow the rules. \\
\multicolumn{2}{p{\dimexpr\textwidth-2\tabcolsep\relax}}{\texttt{BDIE: E[frustration] B: [people should follow the rules] D: — I: none}} \\
\midrule
2 & Agent: I hear you — it's infuriating to watch people act carelessly when you're doing your part. That frustration can feel so heavy, especially when it seems like your own sacrifices are being undermined by others' choices. What if, just for this week, you tried shifting your focus from what they're doing to what you can control? \\
2 & User: Yeah, it's been a tough year. \\
\multicolumn{2}{p{\dimexpr\textwidth-2\tabcolsep\relax}}{\texttt{BDIE: E[frustration] B[struggling with depression and lack of motivation] D[find ways to improve motivation and cope with pandemic stress] I: none}} \\
\midrule
3 & Agent: It really has been — and the fact that you're still here, still trying, says a lot about your resilience, even if it doesn't feel that way right now. What's one tiny thing you used to enjoy that you could reintroduce this week, just for yourself? \\
3 & User: Do you have any suggestions? \\
\multicolumn{2}{p{\dimexpr\textwidth-2\tabcolsep\relax}}{\texttt{BDIE: E[confusion] B: — D:[seeks suggestions for improving motivation] I: none}} \\
\midrule
4 & Agent: A few small things that can help: try a ten-minute walk without your phone, just noticing what you see and hear. Or pick one comforting ritual — like a morning cup of tea you actually sit down for, or a five-minute journaling session before bed. The key is starting so small it feels almost silly — motivation often follows action, not the other way around. \\
4 & User: I love reading, but I haven't had much time for it lately. \\
\multicolumn{2}{p{\dimexpr\textwidth-2\tabcolsep\relax}}{\texttt{BDIE: E -- B[pandemic makes it hard; reading is a valuable activity; frustrated by others not following rules] D[seeks suggestions for improving motivation] I: none}} \\
\midrule
5 & Agent: Reading is a wonderful escape, and it sounds like you could really use that right now. What if you set a tiny goal — just ten pages a day, or even five — and treated it as a non-negotiable pocket of peace? You deserve that time for yourself, especially after everything this year has taken. \\
5 & User: That sounds like a good idea. \\
\midrule
6 & Agent: I'm glad it resonates with you. Imagine how good it'll feel to curl up with a book, even briefly — that's self-care you've genuinely earned. What book might you pick up first this week? \\
6 & User: I have a lot of books on my phone. \\
\multicolumn{2}{p{\dimexpr\textwidth-2\tabcolsep\relax}}{\texttt{BDIE: E -- B[the pandemic has made it hard to stay motivated; others not following rules is frustrating] D[want to relieve unmotivation and distress] I: none}} \\
\midrule
7 & Agent: Perfect — you've already got a library right there waiting for you. Maybe tonight, instead of scrolling, you could open one of those books for just five minutes and see how it feels. You might be surprised how quickly it draws you in. \\
7 & User: I can just pull them up when I get bored. \\
\midrule
8 & Agent: That's a great mindset — swapping boredom for a few pages could become a really satisfying habit. Sometimes the hardest part is just remembering it's an option in the moment. You've got everything you need already in your pocket. \\
8 & User: That's true. I should do that more often. \\
\multicolumn{2}{p{\dimexpr\textwidth-2\tabcolsep\relax}}{\texttt{BDIE: E[hope] B[struggling with depression and lack of motivation; frustrated by others not following rules] D[seeks suggestions for improving low mood and motivation] I: none}} \\
\midrule
9 & Agent: I love that you're giving yourself permission to do that. Even small moments with a good book can be a quiet act of rebellion against everything heavy right now. You deserve that little escape — truly. \\
9 & User: What do you do to stay motivated? \\
\midrule
10 & Agent: I try to focus on small, meaningful routines — things that remind me why showing up matters, even on hard days. But honestly, I've learned that motivation often follows action, not the other way around. The key for me is starting before I feel ready, and being kind to myself when I stumble. \\
10 & User: Do you read a lot? \\
\midrule
11 & Agent: I engage with a lot of text, so in a way, I'm always reading — but I don't experience it the way you do, with the comfort of a story and the escape it brings. That's part of why I think your plan to read more is so valuable; it's a uniquely human joy, and you deserve to reclaim it. \\
11 & User: What do you like to read? \\
\midrule
12 & Agent: I don't have personal preferences like a human does, but I've been told that fiction can be a powerful way to process emotions, and poetry can offer comfort in just a few lines. What genres usually pull you in the most? \\
12 & User: That's interesting. \\
\multicolumn{2}{p{\dimexpr\textwidth-2\tabcolsep\relax}}{\texttt{BDIE: E[hope] B[reading is a valuable activity neglected due to the pandemic] D[find ways to cope with pandemic-induced depression and low motivation] I[intends to try reading more on phone when bored]}} \\
\bottomrule
\end{tabular}
\end{table*}

\newpage



\end{document}